\pdfoutput=1
\documentclass[10pt,twocolumn,letterpaper]{article}
\usepackage[pagenumbers]{cvpr}
\usepackage{cuted}
\usepackage{colortbl}
\usepackage{marvosym}
\definecolor{cvprblue}{rgb}{0.21,0.49,0.74}
\usepackage[pagebackref,breaklinks,colorlinks,allcolors=cvprblue]{hyperref}

\title{SAM 3 Sees More Than It Detects:\\Promptable Concept Maps Beyond Categories}

\author{Yu-Huan Wu$^{1}$\textsuperscript{\Letter} \quad Yun Liu$^{2}$ \quad Liangli Zhen$^{1}$\\
$^{1}$CFAR, A*STAR, Singapore\qquad $^{2}$Nankai University}

\begin{document}
\maketitle
{\let\thefootnote\relax\footnotetext{\textsuperscript{\Letter}Corresponding Author: Yu-Huan Wu (wyh.nku@gmail.com)}}

\begin{strip}
\centering
\includegraphics[width=\textwidth]{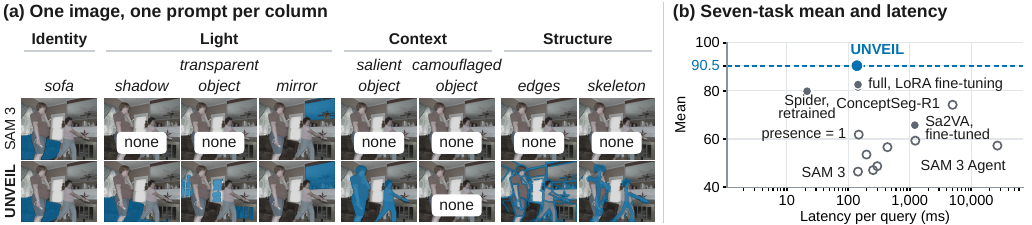}
\captionsetup{hypcap=false}
\captionof{figure}{\textbf{What SAM~3 detects and what UNVEIL returns.}
(a)~One image, one prompt per column.
Native SAM~3 answers ``sofa'' and ``mirror'' only.
UNVEIL answers every prompt with one set of weights, and its presence decision correctly returns no mask for ``camouflaged object''.
Its ``shadow'' mask also covers the dark sofa.
(b)~Seven-task mean of \cref{tab:scorecard} against latency per query.
Filled points: trained on our data.
Hollow points: released weights, unlabeled for LISA-7B, GLaMM, PixelLM, Sa2VA and Grounded SAM~2.}
\captionsetup{hypcap=true}
\label{fig:teaser}
\end{strip}

\begin{abstract}
Promptable segmenters answer what a thing is.
Dense prediction tasks also ask class-agnostic questions, such as what stands out or hides, how light is blocked, transmitted or reflected, and where structure lies.
Robots, surveillance systems and image editors build on their answers.
We study promptable concept maps (PCM), which extend SAM~3's promptable concept segmentation from noun instances to maps of concepts named by short noun phrases such as ``transparent object'', with ``none'' as an answer.
\mbox{PCM-Bench} groups the standard test sets of seven such tasks by the question they ask.
On it, we find that SAM~3 sees more than it detects.
Its semantic path responds to many of these concepts, yet a low image-level presence score can block every detection of its instance path.
We present UNVEIL, which trains a copy of SAM~3's semantic path for these maps, with LoRA on copies of its last blocks.
A separate presence decoder learns from verified negatives whether a concept occurs.
With one text phrase per query, UNVEIL maps region concepts such as salient and camouflaged objects, shadows, transparent objects and mirrors.
The same weights map thin structures such as edges and skeletons as well as object categories.
Run next to SAM~3, it leaves SAM~3's own outputs unchanged.
On each of the seven tasks, UNVEIL reaches the highest score among the compared promptable models, including SAM~3's official fine-tuning and pipelines built on multimodal LLMs.
Code is available at \url{https://github.com/yuhuan-wu/UNVEIL}.
\end{abstract}

\section{Introduction}
\label{sec:intro}

Promptable segmenters let a vision system name what it wants in a short phrase and get its mask back.
Vision agents already call them as tools~\cite{chen2026toward,zhao2026renaissance}, and the phrases they answer name things, such as objects, parts and stuff.
Many dense prediction tasks ask other questions of every pixel.
Which object stands out~\cite{wang2017learning}, and which one hides in its surroundings~\cite{fan2020camouflaged}?
Where does a shadow fall~\cite{vicente2016large}, and where are glass~\cite{xie2020segmenting} and mirrors~\cite{yang2019where}?
Where do the edges~\cite{arbelaez2011contour} and the skeleton~\cite{shen2017deepskeleton} of a shape run?
These questions are class-agnostic, since a salient or camouflaged object can belong to any category and a shadow or an edge belongs to none.
Each question has its own benchmarks and specialist models, and each answer is a map.
Depth sensors misread glass and mirrors, so robots must find them before grasping objects or mapping a room~\cite{sajjan2020cleargrasp,tan2021mirror3d}.
Surveillance separates moving objects from their shadows~\cite{prati2003detecting}, and image editors take a subject, its shadow or its edges as control signals~\cite{zhang2023adding}.
For an absent concept, the right answer is ``none''.

SAM~3~\cite{carion2025sam3} introduced promptable concept segmentation, which returns every instance of a noun phrase.
Unified dense models answer several of the questions above with one network.
Each specifies its task by a learned task prompt~\cite{luo2024vscode,liu2023explicit}, an image--mask example~\cite{zhao2024spider} or a task head~\cite{liu2020dynamic}.
Other systems pass harder queries through a multimodal language model before SAM~3~\cite{zhao2026conceptseg}.
The dense questions call for SAM~3's interface at the level of maps, with one noun phrase in, such as ``shadow'' or ``transparent object'', and one map or ``none'' out.
We refer to this setting as \emph{promptable concept maps} (PCM).
For analysis, we group concepts by the question they ask of a region into four kinds, Identity, Light, Context and Structure.
PCM-Bench is a unified evaluation protocol on the official test splits of seven tasks and on LVIS categories left out of training~\cite{gupta2019lvis}.

\begin{figure}[t]
\centering
\includegraphics[width=\linewidth]{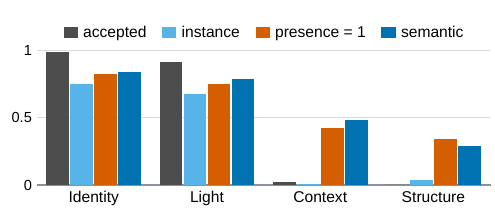}
\caption{\textbf{What frozen SAM~3 accepts and what it sees.}
Medians per kind.
Accepted: share of images with a detection score of at least 0.5.
Other bars score the instance map, the same map with $\sigma(p)=1$ and the semantic map in IoU (ODS for Structure).}
\label{fig:ladder}
\end{figure}

Since SAM~3 already takes a phrase as its prompt, we first run it on PCM-Bench.
On many prompts beyond categories, it returns nothing (\cref{fig:teaser}a).
We find that SAM~3 carries more of these concepts than its detections show.
Its instance path multiplies every detection score by an image-level presence score, and its semantic path predicts a map before this score is applied.
SAM~3 accepts Identity and Light prompts on most images and Context and Structure prompts on almost none (\cref{fig:ladder}).
There the presence score is suppressive, yet the semantic path still holds spatial evidence for the prompt.
Setting the presence score to one, and changing nothing else, restores much of this evidence in the instance path on every Context and Structure set.
We call this suppression the presence veto.
It also depends on wording, since SAM~3 accepts ``glass'' on most Trans10K test images and ``see-through object'' on hardly any of them.

\begin{figure*}[t]
\centering
\includegraphics[width=\textwidth]{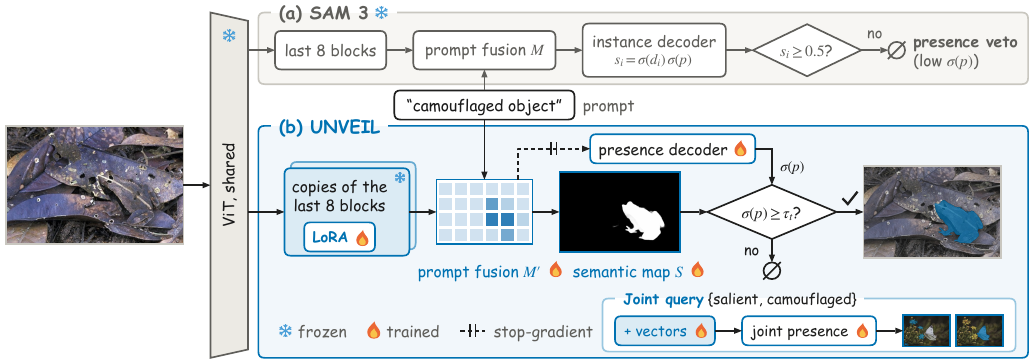}
\caption{\textbf{UNVEIL.}
The image passes the shared, frozen ViT blocks once.
(a)~SAM~3's last blocks and instance decoder, whose presence $\sigma(p)$ scales every detection score $s_i$ and here rejects ``camouflaged object''.
(b)~LoRA-tuned copies of these blocks feed the semantic map $S$.
A presence decoder reads the adapted memory $M'$ through a stop-gradient and keeps $S$ if $\sigma(p)$ reaches the threshold $\tau_t$.
Inset: the joint query \{salient, camouflaged\}, a pair-specific extension.}
\label{fig:overview}
\end{figure*}

We build UNVEIL, short for Uncovering Non-detected Visual Evidence In Latents, on this finding (\cref{fig:overview}).
Where a concept lies and whether it occurs are separate questions, and UNVEIL answers them with separate parts.
It reads the maps from SAM~3's semantic path and trains this path in a parallel branch of LoRA-tuned copies of SAM~3's last blocks.
A presence decoder learns from verified negatives whether a concept occurs, behind a stop-gradient that keeps its loss away from the maps.
Task concepts and category names share one text interface, and the model still segments LVIS categories left out of training.
Run next to SAM~3, UNVEIL leaves SAM~3's weights and outputs unchanged.
Salient and camouflaged objects follow two labeling conventions, one concept at a time or a partition of the image.
For this pair, a learned \emph{convention vector} per prompt selects the partition convention of USC12K~\cite{zhou2025rethinking}, and a \emph{joint query} splits a scene between the two concepts.

On PCM-Bench, UNVEIL scores highest on every task among the compared promptable models (\cref{tab:scorecard}), which range from frozen SAM~3 and its official fine-tuning to SAM~3 Agent and ConceptSeg-R1~\cite{zhao2026conceptseg}.
Its seven-task mean is 90.5, against 82.6 for the next best of them.
SAM~3's official fine-tuning trains its instance path, and it stays below UNVEIL on almost every test set, with all its parameters or with our budget.
The choice of path matters more than the number of trained parameters.

In summary, our main contributions include:
\begin{itemize}
\item \textbf{The presence veto.}
Across the Context and Structure sets, SAM~3's suppressive presence scores coexist with usable spatial predictions of its semantic path.
\item \textbf{UNVEIL.}
Reading the maps from SAM~3's semantic path in a parallel branch, and learning presence apart from them, turns SAM~3 into one promptable model for regions, thin structures and categories.
It scores highest among the compared promptable models on each of the seven tasks, and convention vectors extend it to the salient and camouflaged pair.
\item \textbf{PCM-Bench.}
A unified evaluation protocol for seven class-agnostic tasks on their standard test sets, with a rejection split, a scorecard and a released toolkit.
\end{itemize}

\section{Related Work}
\label{sec:related}

\paragraph{Specialist and unified dense models.}
Each task has strong models trained under its own protocol~\cite{wei2020f3net,wu2022edn,liu2024vst,zheng2024bilateral,chen2026liquid,fan2020camouflaged,fan2022concealed,pang2024zoomnext,xie2015holistically,liu2017richer}, and recent ones build on foundation encoders, diffusion or masked prediction~\cite{han2026beyond,he2026refining,rust2026bedsam2,rust2026resolution,nakamura2026easycontroledge,liufu2025sauge,gao2026amped,cheng2026memo}.
DFI~\cite{liu2020dynamic} predicts salient objects, edges and skeletons with fixed task heads, and MLoRE~\cite{yang2024multi} mixes low-rank experts.
EVP~\cite{liu2023explicit,liu2026explicit}, VSCode~\cite{luo2024vscode} and Spider~\cite{zhao2024spider} select the task with learned or visual prompts, and H2Net~\cite{han2026homo} with task adapters.
FOCUS~\cite{you2025focus} and GateNet~\cite{zhao2024towards} share one architecture across foreground tasks, GateNet with one model per task.
SENet~\cite{hao2025simple} and USCNet~\cite{zhou2025rethinking} train salient and camouflaged detection together, SENet with a shared encoder and two decoders, and USCNet with exclusive outputs.
In-context learners~\cite{wang2023seggpt,reiss2025visual} infer the task from example pairs.
UNVEIL selects the task with a short text prompt and maps regions and thin structures with one head.

\paragraph{Adapting SAM.}
Adapters and U-Net-style decoders tune SAM~\cite{kirillov2023segment}, SAM~2~\cite{ravi2025sam2} and SAM~3 for camouflage, shadows and related tasks~\cite{chen2023sam,xiong2026sam2unet,chen2025sam3adapter,xiong2025sam3unet}, and SAM without adaptation struggles in concealed scenes~\cite{ji2023sam}.
Both SAM~3 adaptations~\cite{chen2025sam3adapter,xiong2025sam3unet} keep its image encoder and replace the text prompt with a decoder for each task.
LoRA~\cite{hu2022lora} learns low-rank updates that merge into weights, and ladder side-tuning~\cite{sung2022lst} trains a side network next to a frozen backbone.
Fine-tuning can distort pretrained features~\cite{kumar2022fine}, and weight ensembling~\cite{wortsman2022robust}, rehearsal~\cite{rolnick2019experience} and distillation~\cite{li2018learning} limit this.
We add LoRA to copies of the last encoder blocks, compare training paths in SAM~3 and replay LVIS masks.

\paragraph{Text-promptable segmentation.}
CLIPSeg~\cite{luddecke2022image} predicts a dense map from a text or image prompt.
Grounding DINO~\cite{liu2024grounding} detects open-set phrases, and Grounded SAM~\cite{ren2024grounded} turns its boxes into masks.
GRES~\cite{liu2023gres} and GSVA~\cite{xia2024gsva} let a referring expression match several targets or none.
UniLSeg~\cite{liu2024universal} segments at many granularities from language instructions, salient objects included.
SAM~3~\cite{carion2025sam3} returns every instance of a noun phrase and adds a presence token that separates recognition from localization.
On counterfactually edited images, CAFE~\cite{liang2026pixels} finds that SAM~3 and other models often return accurate masks for misleading prompts and fail to reject invalid concepts.
The presence veto concerns the opposite error, where SAM~3's presence score rejects present concepts that its semantic path localizes, and setting it to one recovers them.
ConceptSeg-R1~\cite{zhao2026conceptseg} places a multimodal language model in front of frozen SAM~3.
DCR-SAM~\cite{liao2026efficient} and ActiveSAM~\cite{tien2026activesam} adapt SAM~3 to many-class semantic segmentation.
SegEarth-OV3~\cite{li2025segearth} fuses SAM~3's semantic and instance heads without training and drops absent classes of remote-sensing images by the presence score.
UNVEIL trains the semantic path for concepts beyond categories and thin structures, and learns presence from verified negatives.

\section{Concept Maps and the Presence Veto}
\label{sec:concepts}

\subsection{Promptable Concept Maps}
\label{sec:task}

A query pairs an image with a short noun phrase $t$, such as ``mirror'' or ``edges'', in the form of SAM~3's prompts.
One model answers the queries of every task.
The answer is the map of the concept that $t$ names, or ``none'' if the image does not show it.
The map marks a region, as for a shadow, or a thin structure, as for edges.
Promptable concept segmentation~\cite{carion2025sam3} returns the instances of a noun phrase, and referring segmentation~\cite{liu2023gres,xia2024gsva} the instances that an expression describes.
Open-vocabulary semantic segmentation~\cite{li2025segearth,tien2026activesam} labels pixels with categories, and unified dense models~\cite{luo2024vscode,zhao2024spider} specify each task by a task token or a visual prompt.
Categories are one kind of concept among several in PCM.
PCM-Bench queries each test set with the phrase of its training split, tests rewordings for robustness and scores ``none'' where negatives are verified.
Each test set keeps its standard metric.
Shadows use the balanced error rate (BER), glass and mirrors IoU, and salient and camouflaged objects the S-measure $S_\alpha$~\cite{fan2017structure}.
Edges and skeletons use ODS, the F-measure at the best threshold over a test set.

Concepts differ in what they ask of a region.
\emph{Identity} asks what a thing is, as for objects, parts and stuff.
\emph{Light} asks how light meets a region, as for shadows, glass and mirrors.
\emph{Context} asks how a region relates to its scene or viewer, as for salient and camouflaged objects.
\emph{Structure} asks where the geometry of a shape lies, as for edges and skeletons.
PCM-Bench covers Identity with LVIS categories and the other kinds with seven tasks.

\subsection{Two output paths in SAM~3}
\label{sec:method-prelim}

SAM~3~\cite{carion2025sam3} encodes a $1008\times1008$ image with a 32-block Perception Encoder ViT~\cite{bolya2025perception} and a noun-phrase prompt with a text encoder.
A fusion encoder lets image tokens attend to prompt tokens and yields a prompt-conditioned memory $M$ on a $72\times72$ grid, which two heads decode.

The \emph{instance path} is a DETR-style decoder~\cite{carion2020end} with 200 object queries and one presence token.
Query $i$ predicts a box, a mask and a detection logit $d_i$.
The presence token predicts one logit $p$ for the whole image.
With $\sigma$ the sigmoid, a detection scores $s_i=\sigma(d_i)\,\sigma(p)$, and SAM~3 keeps the detections with $s_i\ge0.5$, so a low $\sigma(p)$ suppresses every detection.
To score this path as a map, we take per pixel the highest $s_i$ times the mask over the detections with $s_i\ge0.05$, the \emph{instance map}.
The \emph{semantic path} attends from $M$ to the prompt once more.
A small pixel decoder and a $1\times1$ convolution then give a logit map $S$ of $288\times288$ pixels.
SAM~3 pairs $S$ with the same presence score for semantic segmentation~\cite{carion2025sam3}, and we read $S$ before this score is applied.

\subsection{Where the veto falls}
\label{sec:veto}

We run frozen SAM~3 on every image of the 15 test sets of PCM-Bench and on Identity concepts from LVIS~\cite{gupta2019lvis}, ADE20K~\cite{zhou2017scene} and SA-Co/Gold~\cite{carion2025sam3}.
The acceptance rate is the share of images with a detection $s_i\ge0.5$.
Its median is 0.99 for Identity, 0.91 for Light, 0.02 for Context and 0.00 for Structure (\cref{fig:ladder}), so the veto depends on the kind of concept.
SAM~3.1 has medians of 0.99, 0.92, 0.02 and 0.00, and across the 15 test sets its acceptance rates follow those of SAM~3 with a Spearman correlation of 0.99.
We examine the kinds and wordings where acceptance collapses.

\begin{table}[t]
\centering\footnotesize
\setlength{\tabcolsep}{5pt}
\begin{tabular}{@{}lcccc@{}}
\toprule
 & & \multicolumn{2}{c}{Instance map} & \\
\cmidrule(lr){3-4}
Test set & Accepted & as is & $\sigma(p)=1$ & Semantic map \\
\midrule
\multicolumn{5}{@{}l}{\emph{Context, IoU}} \\
DUTS-TE & 0.02 & 0.002 & 0.398 & 0.424 \\
ECSSD & 0.02 & 0.009 & 0.410 & 0.483 \\
HKU-IS & 0.01 & 0.008 & 0.430 & 0.508 \\
DUT-OMRON & 0.02 & 0.002 & 0.258 & 0.269 \\
PASCAL-S & 0.03 & 0.015 & 0.188 & 0.249 \\
CAMO & 0.12 & 0.066 & 0.461 & 0.492 \\
COD10K & 0.02 & 0.010 & 0.469 & 0.473 \\
NC4K & 0.04 & 0.023 & 0.517 & 0.522 \\
\midrule
\multicolumn{5}{@{}l}{\emph{Structure, ODS}} \\
BSDS500 & 0.00 & 0.035 & 0.340 & 0.309 \\
SK-LARGE & 0.00 & 0.234 & 0.533 & 0.287 \\
SYM-PASCAL & 0.00 & 0.000 & 0.190 & 0.142 \\
\bottomrule
\end{tabular}
\caption{\textbf{The presence veto per test set.}
Frozen SAM~3 with each set's training prompt, accepted as in \cref{fig:ladder}.
The two instance columns differ only in $\sigma(p)$.
The instance map keeps detections with $s_i\ge0.05$, so ODS is nonzero where SAM~3 accepts nothing.}
\label{tab:veto-main}
\end{table}

On COD10K, SAM~3 accepts ``camouflaged object'' on 0.02 of the images, so its presence score suppresses almost every detection.
Its instance map reaches an IoU of 0.010.
The semantic path holds spatial evidence for the same prompt and reaches 0.473.
We score these maps by IoU after binarizing them at 0.5, since an empty map still receives an S-measure near 0.5.
A rejected image then has an empty instance map.
SAM~3 itself sets the presence score to one when the target is known to be present, as for visual prompts on a single frame or for counting~\cite{carion2025sam3}.
We set $\sigma(p)=1$ and change nothing else, which lifts the instance path to 0.469.
Structure shows the same pattern on BSDS500, where SAM~3 accepts ``edges'' on no image.
Setting $\sigma(p)=1$ lifts the instance map there from an ODS of 0.035 to 0.340, above the semantic path's 0.309.
On the Context sets, this intervention recovers 74\% to 99\% of the IoU gap between the paths, and on the Structure sets it exceeds the semantic map in ODS (\cref{tab:veto-main}).
The semantic path finds the region from the same text embedding, so the veto is not a failure to encode the phrase.

The wording alone can trigger the veto.
On Trans10K, SAM~3 accepts ``glass'' on 0.93 of the test images and ``see-through object'' on 0.01, yet the semantic map reaches an IoU of 0.878 and 0.840 for the two prompts.
On COD10K, it accepts the class name of each animal on 0.67 of the images and ``camouflaged animal'' on 0.06, and the semantic map reaches 0.797 and 0.767.
The noun passes more often in all 18 noun--phrase pairs we score.

SAM~3 applies the same presence score to its semantic output, which suppresses that output as well.
\Cref{tab:veto-main} reports the semantic map without this score.
Setting presence to one closes much of the gap, but it gives up the answer ``none'' for absent concepts, and the instance path returns binary masks per object, which fit thin structures poorly.
UNVEIL therefore trains the semantic path and lets presence decide only whether a concept occurs (\cref{sec:exp-path}).

\section{UNVEIL}
\label{sec:method}

UNVEIL trains copies of SAM~3's fusion encoder, semantic path, presence decoder and last blocks for PCM (\cref{fig:overview}), and each task is a text prompt such as ``shadow'' or ``edges''.
LoRA~\cite{hu2022lora} and the semantic head are existing parts, and we choose where to read the maps and how to learn presence.

\subsection{Training the semantic path}
\label{sec:method-model}

UNVEIL trains the semantic path together with the fusion encoder, the presence decoder and a LoRA branch, and keeps the ViT and the text encoder frozen.
The LoRA branch and the trained fusion encoder give an adapted memory $M'$.
Each task supervises $\sigma(S)$ with its standard loss.
Region tasks use BCE plus an IoU loss, with class-balanced BCE for shadows.
Edges and skeletons use the class-balanced BCE of RCF~\cite{liu2017richer} and HED~\cite{xie2015holistically}.
The dense loss averages the per-task means.

\paragraph{Shared LoRA branch.}
We copy SAM~3's last ViT blocks and add LoRA~\cite{hu2022lora} to their attention and MLP layers.
The same branch serves all tasks without task-specific weights, and its adapters merge into the copies after training.

\paragraph{Two deployments.}
When only our maps are needed, the copies replace SAM~3's last blocks, and one semantic map replaces the per-detection masks that SAM~3 upsamples and merges.
Otherwise they run in parallel with SAM~3's own last blocks, which still feed its native heads, so SAM~3 keeps its weights and outputs.

\paragraph{Concept replay.}
Training on seven tasks erodes the open vocabulary of the semantic path, so we replay Identity concepts.
A replay sample pairs an LVIS image~\cite{gupta2019lvis} with a category name, and the union of that category's instance masks is its target under the region loss.
Replay leaves out the categories and images used to measure retention, and categories related to glass and mirrors.

\subsection{Presence apart from the maps}
\label{sec:method-presence}

The presence decoder, a copy of SAM~3's instance decoder whose presence token gives the logit, decides whether a concept occurs.
It is trained with class-balanced BCE on this logit.
Its positives are the training images of all tasks, and its negatives are images verified to lack the concept.
They include COD10K background and non-camouflaged images~\cite{fan2020camouflaged} for ``camouflaged object'', USC12K scenes with neither concept~\cite{zhou2025rethinking} for both Context prompts, and absent categories for replayed LVIS concepts.

The decoder reads the adapted memory $M'$ and the prompt tokens through a stop-gradient, so its loss never reaches the fusion encoder or the LoRA branch.
The dense loss alone shapes the shared features, and without the stop-gradient 8 of the 15 test sets score lower (\cref{sec:exp-ablation}).
A single query with prompt $t$ returns $\sigma(S)$ at image size if $\sigma(p)\ge\tau_t$, else an empty map.
The threshold $\tau_t$, fixed before testing, maximizes balanced accuracy on validation pairs from the training splits.
Each prompt takes one forward pass, without a language model or a reference image.

\subsection{An extension for two conventions}
\label{sec:method-joint}

Salient and camouflaged objects come with two labeling conventions.
Both concepts depend on the scene, so the same kind of object can be either, as a butterfly is salient in ECSSD and camouflaged in COD10K.
Single-concept benchmarks label each concept alone, and USC12K~\cite{zhou2025rethinking} labels the two as a partition of the image, so the same plain prompt receives different targets in training.
For this pair, a learned \emph{convention vector} per prompt is added to the text tokens of USC12K's training pairs, and single queries never use it.
A \emph{joint query} \{salient object, camouflaged object\} adds the vectors to both members.
A joint presence module behind a stop-gradient then keeps each member that passes a joint threshold $\tau_J$, also the value with the highest balanced accuracy on validation pairs from the training splits.
We show this convention-conditioned extension for the salient and camouflaged pair only.

\section{Experiments}
\label{sec:experiments}

\begin{table}[t]
\centering\scriptsize
\setlength{\tabcolsep}{2.4pt}
\begin{tabular}{@{}l>{\raggedright\arraybackslash}p{4.75cm}rl@{}}
\toprule
Kind & \emph{Prompt}: test sets & Queries & Metric \\
\midrule
Identity & \emph{LVIS category}: 200 categories left out of replay & 2,294 & IoU \\
\midrule
Light & \emph{shadow}: SBU, ISTD & 1,178 & $100-$BER \\
 & \emph{transparent object}: Trans10K & 4,428 & IoU \\
 & \emph{mirror}: MSD & 955 & IoU \\
\midrule
Context & \emph{salient object}: DUTS-TE, ECSSD, HKU-IS, DUT-OMRON, PASCAL-S & 16,484 & $S_\alpha$ \\
 & \emph{camouflaged object}: CAMO, COD10K, NC4K & 6,397 & $S_\alpha$ \\
\midrule
Structure & \emph{edges}: BSDS500 & 200 & ODS \\
 & \emph{skeleton}: SK-LARGE, SYM-PASCAL & 1,533 & ODS \\
\midrule
Rejection & \emph{camouflaged object}: COD10K background & 784 & ``none''$\uparrow$ \\
 & \emph{salient / camouflaged object}: USC12K without the concept & 3,600 & ``none''$\uparrow$ \\
 & \emph{LVIS category}: verified absent from the image & 400 & ``none''$\uparrow$ \\
\bottomrule
\end{tabular}
\caption{\textbf{PCM-Bench.}
Kinds in the order of SAM~3's acceptance (\cref{sec:veto}).
Rejection: share of queries answered with no map.}
\label{tab:bench}
\end{table}
\begin{table*}[t]
\centering\scriptsize
\setlength{\tabcolsep}{1.35pt}
\begin{tabular}{lcccccccccccccccc}
\toprule
 & \multicolumn{4}{c}{Light} & \multicolumn{8}{c}{Context} & \multicolumn{3}{c}{Structure} &  \\
\cmidrule(lr){2-5}\cmidrule(lr){6-13}\cmidrule(lr){14-16}
 & \multicolumn{2}{c}{Shadow} & Transp. & Mirror & \multicolumn{5}{c}{Salient} & \multicolumn{3}{c}{Camouflaged} & Edge & \multicolumn{2}{c}{Skeleton} & \\
\cmidrule(lr){2-3}\cmidrule(lr){4-4}\cmidrule(lr){5-5}\cmidrule(lr){6-10}\cmidrule(lr){11-13}\cmidrule(lr){14-14}\cmidrule(lr){15-16}
 & SBU & ISTD & Trans10K & MSD & DUTS-TE & ECSSD & HKU-IS & DUT-O & PASC.-S & CAMO & COD10K & NC4K & BSDS500 & SK-LARGE & SYM-P. & Mean \\
 & BER\,$\downarrow$ & BER\,$\downarrow$ & IoU & IoU & $S_\alpha$ & $S_\alpha$ & $S_\alpha$ & $S_\alpha$ & $S_\alpha$ & $S_\alpha$ & $S_\alpha$ & $S_\alpha$ & ODS & ODS & ODS & \\
\midrule
SAM~3 (frozen) & 12.62 & 10.25 & 0.585 & 0.692 & 0.437 & 0.405 & 0.441 & 0.440 & 0.402 & 0.519 & 0.503 & 0.517 & 0.035 & 0.233 & 0.000 & 46.5 \\
SAM~3, presence set to one & 12.39 & 10.57 & 0.806 & 0.769 & 0.645 & 0.712 & 0.747 & 0.597 & 0.601 & 0.504 & 0.483 & 0.535 & 0.340 & 0.532 & 0.190 & 61.8 \\
SAM~3 Agent (7B)~\cite{carion2025sam3} & 22.11 & 17.51 & 0.697 & 0.727 & 0.835 & 0.878 & 0.814 & 0.791 & 0.821 & 0.742 & 0.806 & 0.834 & 0.086 & 0.041 & 0.109 & 57.3 \\
\midrule
LISA-7B~\cite{lai2024lisa} & 34.99 & 31.12 & 0.533 & 0.529 & 0.707 & 0.787 & 0.812 & 0.657 & 0.719 & 0.678 & 0.716 & 0.694 & 0.126 & 0.229 & 0.013 & 48.7 \\
GLaMM~\cite{rasheed2024glamm} & 32.20 & 24.89 & 0.588 & 0.619 & 0.767 & 0.867 & 0.764 & 0.743 & 0.810 & 0.746 & 0.704 & 0.698 & 0.214 & 0.389 & 0.255 & 56.6 \\
PixelLM-7B~\cite{ren2024pixellm} & 37.38 & 33.92 & 0.589 & 0.617 & 0.759 & 0.849 & 0.750 & 0.716 & 0.789 & 0.712 & 0.701 & 0.746 & 0.180 & 0.347 & 0.107 & 53.5 \\
Grounded SAM~2~\cite{ren2024grounded} & 34.95 & 28.64 & 0.456 & 0.240 & 0.756 & 0.802 & 0.756 & 0.661 & 0.655 & 0.481 & 0.570 & 0.636 & 0.125 & 0.571 & 0.439 & 47.1 \\
Sa2VA-8B~\cite{yuan2026sa2va} & 27.81 & 20.13 & 0.721 & 0.563 & 0.832 & 0.879 & 0.747 & 0.816 & 0.820 & 0.765 & 0.759 & 0.743 & 0.124 & 0.581 & 0.238 & 59.3 \\
ConceptSeg-R1~\cite{zhao2026conceptseg} & 11.92 & 7.77 & 0.886 & 0.831 & 0.937 & 0.952 & 0.935 & 0.877 & \textbf{0.900} & 0.849 & 0.895 & 0.902 & 0.323 & 0.554 & 0.348 & 74.2 \\
\midrule
DFI~\cite{liu2020dynamic} & -- & -- & -- & -- & 0.887 & 0.927 & 0.920 & 0.840 & 0.865 & -- & -- & -- & 0.799 & 0.723 & 0.511 & -- \\
VSCode~\cite{luo2024vscode} & -- & -- & -- & -- & 0.926 & 0.949 & 0.940 & 0.877 & 0.886 & 0.873 & 0.869 & 0.891 & -- & -- & -- & -- \\
SENet (joint, P2)~\cite{hao2025simple} & -- & -- & -- & -- & 0.929 & 0.950 & 0.940 & 0.874 & 0.891 & 0.861 & 0.849 & 0.878 & -- & -- & -- & -- \\
H2Net~\cite{han2026homo} & -- & -- & -- & -- & 0.939 & -- & -- & -- & -- & -- & 0.891 & -- & -- & -- & -- & -- \\
VisionLLM~v2~\cite{wu2024visionllm} & -- & -- & -- & -- & 0.921 & 0.950 & 0.941 & 0.882 & 0.892 & 0.856 & 0.877 & -- & -- & -- & -- & -- \\
Seg-R1 (3B)~\cite{you2025segr1} & -- & -- & -- & -- & 0.866 & 0.882 & 0.772 & 0.852 & -- & 0.805 & 0.857 & -- & -- & -- & -- & -- \\
Spider~\cite{zhao2024spider} & 3.11 & 3.94 & 0.933 & 0.578 & 0.917 & 0.920 & 0.899 & 0.857 & 0.815 & 0.868 & 0.871 & 0.892 & 0.216 & 0.642 & 0.407 & 71.1 \\
\midrule
Sa2VA-8B~\cite{yuan2026sa2va}, fine-tuned & 9.14 & 9.31 & 0.864 & 0.878 & 0.920 & 0.945 & 0.928 & 0.867 & 0.878 & 0.869 & 0.886 & 0.893 & 0.086 & 0.118 & 0.037 & 65.7 \\
Spider~\cite{zhao2024spider}, retrained & 4.00 & 1.35 & 0.935 & 0.877 & 0.902 & 0.837 & 0.854 & 0.843 & 0.709 & 0.878 & 0.877 & 0.895 & 0.483 & 0.759 & 0.460 & 79.9 \\
SAM~3 full fine-tuning & 4.02 & 1.51 & 0.951 & 0.960 & 0.915 & 0.939 & 0.932 & 0.843 & 0.880 & 0.901 & 0.886 & 0.901 & 0.706 & 0.317 & 0.327 & 81.6 \\
SAM~3 LoRA fine-tuning & 4.40 & 1.74 & 0.950 & 0.955 & 0.923 & 0.900 & 0.906 & 0.850 & 0.861 & 0.899 & 0.907 & 0.893 & 0.634 & 0.550 & 0.420 & 82.6 \\
\midrule
UNVEIL (ours) & \textbf{2.83} & \textbf{0.84} & \textbf{0.956} & \textbf{0.964} & \textbf{0.947} & \textbf{0.956} & \textbf{0.944} & \textbf{0.901} & 0.897 & \textbf{0.918} & \textbf{0.933} & \textbf{0.927} & \textbf{0.856} & \textbf{0.793} & \textbf{0.647} & \textbf{90.5} \\
\bottomrule
\end{tabular}
\caption{\textbf{PCM-Bench scorecard.}
One primary metric per test set.
Blocks: frozen SAM~3 and SAM~3 Agent, text-promptable models on every prompt, multi-task models, models trained on our data, and UNVEIL.
Mean: of the seven task scores.
Task score: the mean primary metric over the task's test sets, $\times100$, with $100-\mathrm{BER}$ for shadows. --: not reported.
DUT-O: DUT-OMRON.
SYM-P.: SYM-PASCAL.
PASC.-S: PASCAL-S.
Bold: best.}
\label{tab:scorecard}
\end{table*}

\subsection{PCM-Bench and setup}
\label{sec:exp-setup}

\paragraph{PCM-Bench.}
The benchmark packages 15 standard test sets of seven tasks~\cite{vicente2016large,wang2018stacked,xie2020segmenting,yang2019where,wang2017learning,yan2013hierarchical,li2015visual,yang2013saliency,li2014secrets,le2019anabranch,fan2020camouflaged,lv2021simultaneously,arbelaez2011contour,shen2017deepskeleton,ke2017srn} with 31,175 images (\cref{tab:bench}).
Identity is scored for retention on LVIS categories left out of replay~\cite{gupta2019lvis} and is no part of the seven tasks or their mean.
Each test image shows its concept, so these sets score the maps of UNVEIL before its presence decision.
A rejection split scores the answer ``none'' on images verified to lack the prompted concept (\cref{tab:bench}).
A task score averages the primary metric over the task's test sets, $\times100$, with $100-\mathrm{BER}$ for shadows.

\paragraph{Data.}
One model learns all seven tasks from their official training splits.
Salient objects come from DUTS-TR~\cite{wang2017learning}, and camouflaged objects from COD10K-TR~\cite{fan2020camouflaged} and CAMO-TR~\cite{le2019anabranch}.
Shadows use SBU~\cite{vicente2016large} and ISTD~\cite{wang2018stacked}, transparent objects Trans10K~\cite{xie2020segmenting}, and mirrors MSD~\cite{yang2019where}.
Edges use BSDS500~\cite{arbelaez2011contour} with HED's augmentation~\cite{xie2015holistically}, and skeletons use SK-LARGE~\cite{shen2017deepskeleton} with the prompt ``skeleton'' and SYM-PASCAL~\cite{ke2017srn} with ``symmetry axis''.
USC12K~\cite{zhou2025rethinking} adds training scenes with both Context concepts, and LVIS~\cite{gupta2019lvis} the replay images.

\paragraph{Protocol.}
Every number comes from the full test set at a single scale, without test-time augmentation or CRF, and tables give BER in \%.
Thresholds for BER and IoU come from the training splits.
Every text-promptable model gets our training prompts, and external models with released weights keep their own thresholds.
Each model is trained once and tested at its last checkpoint.
Two models differ on a test set when the 95\% interval of a paired bootstrap over its images excludes zero (\cref{sec:app-train}).

\paragraph{Compared models.}
The scorecard holds promptable models that answer every prompt with one set of weights.
Frozen SAM~3 and its fine-tuned versions are scored by their instance maps (\cref{sec:method-prelim}).
Frozen SAM~3 runs once as released and once with its presence score set to one.
SAM~3 Agent~\cite{carion2025sam3} runs with Qwen2.5-VL-7B~\cite{bai2025qwen25vl} as its language model.
SAM~3's official full fine-tuning trains every parameter through the instance path, with our data, batches and iterations.
Its LoRA fine-tuning puts our LoRA branch on SAM~3's own blocks and trains the same path.
ConceptSeg-R1~\cite{zhao2026conceptseg} runs with its released weights, and unified models enter with the printed values of their papers or with their released weights under our scorer.

\paragraph{Implementation details.}
The LoRA branch copies ViT blocks 24 to 31, the last eight of 32, with LoRA of rank 16 and scaling factor $\alpha=32$.
The trained modules are the fusion encoder with the semantic path (10.9M), the presence decoder (11.4M) and the LoRA branch (2.26M).
Together they hold 24.6M parameters, or 2.9\% of SAM~3.
The convention vectors add 512 parameters and the joint presence module 11.4M.
UNVEIL trains for 11,000 iterations on 8 RTX 4080 SUPER GPUs, which takes 17.2 hours, with the optimizer settings of \cref{sec:app-train}.
Each step draws 16 images for every task, for the USC12K pairs, for each pool of presence negatives and for replay.
The base model of the ablations omits the joint presence module.

\subsection{One model for every prompt}
\label{sec:exp-main}

\begin{figure*}[t]
\centering
\includegraphics[width=\textwidth]{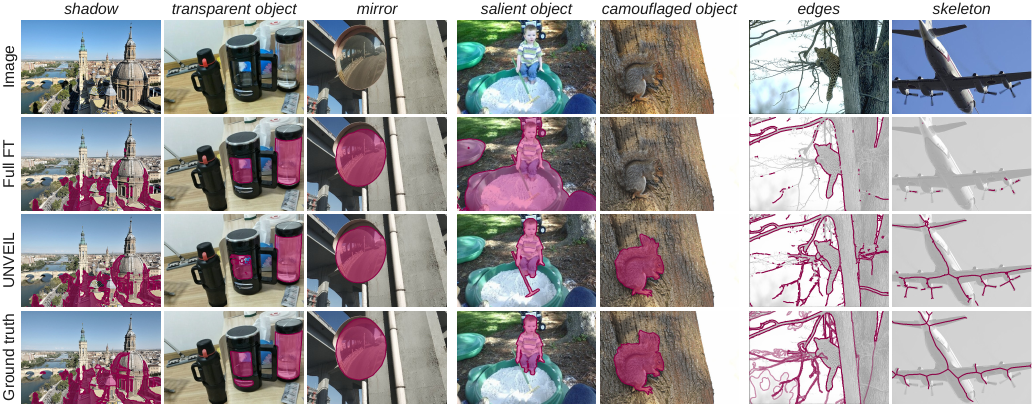}
\caption{\textbf{Examples for the seven prompts.}
Rows show the image, the official full fine-tuning of SAM~3 (Full FT), UNVEIL and the ground truth.
\Cref{fig:qual} shows random test images.}
\label{fig:qual-main}
\end{figure*}

In the PCM-Bench scorecard (\cref{tab:scorecard}), frozen SAM~3 reaches a seven-task mean of 46.5, and 61.8 with $\sigma(p)=1$ and nothing trained.
SAM~3 Agent, whose language model tends to rephrase each prompt as an object name, reaches 57.3, and ConceptSeg-R1 74.2.
Trained on our data, Sa2VA reaches 65.7 with 45.5M trained parameters and Spider 79.9 with 198.3M.
SAM~3's official full and LoRA fine-tuning reach 81.6 and 82.6 with 838.5M and 26.2M, and UNVEIL 90.5 with 24.6M.
Setting presence to one changes the two fine-tuned rows little, to 81.8 and 83.0.
UNVEIL has the highest score on every task.
It leads on 14 of the 15 test sets, and \cref{fig:qual-main} shows examples.
UNVEIL also scores highest among the promptable models on UCF, GSD, PMD, CHAMELEON, BIPEDv2 and WH-SYMMAX, six more standard test sets of these tasks (\cref{tab:extrasets}).
Ranked with the task-specific models of \cref{sec:app-compare} by the mean rank over each task's test sets, it is first or tied first on six tasks and second on salient objects.
Against prediction maps of salient and camouflaged models, the paired interval favors UNVEIL in S-measure on 85 of 100 pairs of model and test set and the other on 2.

\begin{table}[t]
\centering\scriptsize
\setlength{\tabcolsep}{1pt}
\begin{tabular*}{\linewidth}{@{\extracolsep{\fill}}lcccccccccr@{}}
\toprule
 & Shad. & Trans. & Mirr. & Sal. & Cam. & Edge & Skel. & Mean & Trained \\
\midrule
On SAM~3 & \textbf{98.2} & \textbf{95.6} & \textbf{96.4} & 92.9 & \textbf{92.6} & 85.6 & 72.0 & 90.5 & 24.6M \\
On SAM~3.1 & \textbf{98.2} & \textbf{95.6} & \textbf{96.4} & 92.9 & 92.5 & 85.5 & 72.1 & 90.4 & 24.6M \\
Seven versions & 97.8 & \textbf{95.6} & 95.2 & \textbf{93.1} & 92.2 & \textbf{85.9} & \textbf{74.7} & \textbf{90.6} & 76.4M \\
\bottomrule
\end{tabular*}
\caption{\textbf{Versions of UNVEIL.}
Task scores as in \cref{tab:scorecard}.
Seven versions: one model per task, each with a frozen encoder.}
\label{tab:versions}
\end{table}

Trained with the same recipe on SAM~3.1, UNVEIL reaches nearly the same task scores (\cref{tab:versions}).
Seven per-task versions of UNVEIL with frozen encoders reach about the same mean with three times the trained parameters.

\paragraph{Cost.}
In the deployment that returns only maps, a query of UNVEIL takes 138\,ms on one GPU, against 144\,ms for frozen SAM~3, 26.7\,s for SAM~3 Agent and 4.98\,s for ConceptSeg-R1 (\cref{fig:teaser}b).
The image passes the encoder and the copies once, so further prompts add only the per-prompt modules.
In network time, PCM-Bench's eight prompts take 232\,ms in one batch against 129\,ms for one.

\paragraph{Identity.}
On the 200 LVIS categories left out of replay, UNVEIL answers category names through the same text interface, with a mean IoU of 0.841 against 0.831 for frozen SAM~3.
In the parallel deployment, SAM~3 returns its native outputs bit for bit on all 168,202 pairs of SA-Co/Gold~\cite{carion2025sam3}.
Since SAM~3's weights are untouched, its cgF1, the official SA-Co/Gold score, stays at 53.92.
Fine-tuning SAM~3 itself on our data lowers it to 48.67 with full, 45.79 with LoRA and 45.25 with semantic-path fine-tuning.
Full and LoRA fine-tuning lose mostly in IL\_MCC, the benchmark's score for whether the concept is in the image, which falls from 0.815 to 0.740 and 0.707.

\paragraph{Prompt wording.}
We reword three training prompts in two or three ways, fixed before testing.
On DUTS-TE, ECSSD, COD10K and CAMO, UNVEIL reaches an S-measure between 0.871 and 0.956 with any of the prompts, and frozen SAM~3's semantic map between 0.628 and 0.877.
On COD10K, ``camouflaged animal'' reaches 0.931 against 0.933 for the training prompt.
On every reworded test set, the worst prompt of UNVEIL still scores above the best prompt of frozen SAM~3's semantic map.

\subsection{Which path to train}
\label{sec:exp-path}

\begin{table}[t]
\centering\scriptsize
\setlength{\tabcolsep}{1pt}
\begin{tabular*}{\linewidth}{@{\extracolsep{\fill}}llrcccc@{}}
\toprule
 & Path & Trained & Sal. & Cam. & Edge & Mean \\
\midrule
Three tasks & Instance & 11.8M & 87.8 & 86.8 & 55.9 & -- \\
 & Semantic & 10.9M & \textbf{93.1} & \textbf{92.2} & \textbf{85.6} & -- \\
\midrule
LoRA & Instance & 26.2M & 88.8 & 89.9 & 63.4 & 82.6 \\
UNVEIL & Semantic & 24.6M & \textbf{92.9} & \textbf{92.6} & \textbf{85.6} & \textbf{90.5} \\
\midrule
Full & Instance & 838.5M & 90.2 & 89.6 & 70.6 & 81.6 \\
 & Semantic & 837.0M & \textbf{92.7} & \textbf{92.2} & \textbf{86.0} & \textbf{87.1} \\
\bottomrule
\end{tabular*}
\caption{\textbf{Which path to train.}
Task scores as in \cref{tab:scorecard} at three budgets of trained parameters.
Bold: best per budget.}
\label{tab:path}
\end{table}

The path matters more than the budget.
\Cref{tab:path} trains both paths at three budgets with the same data, batches and iterations per pair, and the semantic path scores higher at each.
Training all parameters raises the mean of neither path.
In the three-task study, the gap is largest on small objects.
On salient images whose object covers less than 5\% of the image, the instance path scores 0.189 lower in S-measure, against 0.018 for objects above 30\%.
On BSDS500, it marks 1.4\% of the pixels as edges, against 6.6\% in the annotations.
On edges and skeletons, part of the gap reflects the binary masks of the instance path.

\subsection{Saying ``none'' and two conventions}
\label{sec:exp-context}

\paragraph{Saying ``none''.}
A single query for ``camouflaged object'' is accepted on 2.6\% of the COD10K background images, which contain no camouflaged object.
Presence separates these images and USC12K scenes with neither Context concept from present ones with an AUROC of 0.994.
For LVIS categories verified absent from an image, it declines 98.0\% of the queries.
Without this decision, the maps of UNVEIL return a region for most absent concepts (\cref{tab:reject}).
On the rejection split, the released multimodal models and Spider answer ``none'' on at most 14.2\% of the queries on average (\cref{tab:reject}).
On COD10K backgrounds and absent LVIS categories, the single query of UNVEIL declines about as often as SAM~3's fine-tuned versions.
On USC12K, whether a lone animal is salient depends on the labeling convention~\cite{zhou2025rethinking}, and the joint query, which follows USC12K's convention, declines 95.6\% of these queries.

\begin{table}[t]
\centering\small
\begin{tabular*}{\linewidth}{@{\extracolsep{\fill}}lcc@{}}
\toprule
 & Answers & Says ``none'' \\
 & when present\,$\uparrow$ & when absent\,$\uparrow$ \\
\midrule
SAM~3 (frozen) & 35.9 & 99.6 \\
SAM~3, presence set to one & 90.5 & 22.4 \\
SAM~3 Agent (7B)~\cite{carion2025sam3} & 76.9 & 60.3 \\
\midrule
LISA-7B~\cite{lai2024lisa} & 100.0 & 8.1 \\
GLaMM~\cite{rasheed2024glamm} & 100.0 & 2.1 \\
PixelLM-7B~\cite{ren2024pixellm} & 100.0 & 3.3 \\
Grounded SAM~2~\cite{ren2024grounded} & 82.8 & 38.4 \\
Sa2VA-8B~\cite{yuan2026sa2va} & 99.8 & 14.2 \\
ConceptSeg-R1~\cite{zhao2026conceptseg} & 100.0 & 0.0 \\
\midrule
Sa2VA-8B~\cite{yuan2026sa2va}, fine-tuned & 99.8 & 18.2 \\
Spider~\cite{zhao2024spider}, retrained & 100.0 & 0.0 \\
SAM~3 full fine-tuning & 98.2 & 86.7 \\
SAM~3 LoRA fine-tuning & 98.2 & 89.2 \\
\midrule
UNVEIL, maps only & 100.0 & 9.9 \\
UNVEIL, single query & 98.8 & 84.7 \\
UNVEIL, joint query & 94.3 & 97.5 \\
\bottomrule
\end{tabular*}
\caption{\textbf{Presence decisions.}
Share of present concepts answered and of absent ones declined, in \%, as means over the seven tasks and over the three rejection sources of \cref{tab:bench}.
The joint query of UNVEIL covers the Context pair, with single queries elsewhere.
Maps only: UNVEIL without its presence decision.}
\label{tab:reject}
\end{table}

\begin{table}[t]
\centering\scriptsize
\setlength{\tabcolsep}{1.9pt}
\begin{tabular}{@{}lccccccccc@{}}
\toprule
 & \multicolumn{4}{c}{Accuracy} & Vocab. & \multicolumn{2}{c}{Conventions} & \multicolumn{2}{c}{Rejection} \\
\cmidrule(lr){2-5}\cmidrule(lr){6-6}\cmidrule(lr){7-8}\cmidrule(lr){9-10}
 & \multicolumn{2}{c}{$S_\alpha$} & BER$\downarrow$ & ODS & IoU & \multicolumn{2}{c}{IoU} & \multicolumn{2}{c}{Accept.$\downarrow$} \\
 & Sal. & Cam. & ISTD & SK-L. & LVIS & Sal. & Cam. & Sal. & Cam. \\
\midrule
Base & 0.929 & 0.926 & 0.77 & 0.789 & 0.839 & 0.845 & 0.691 & 0.61 & 0.77 \\
\midrule
w/o LoRA branch & 0.930 & 0.922 & \cellcolor[gray]{0.88}1.23 & \cellcolor[gray]{0.88}0.760 & 0.842 & 0.846 & 0.703 & 0.85 & 0.67 \\
w/o replay & 0.929 & 0.925 & 0.93 & 0.791 & \cellcolor[gray]{0.88}0.728 & 0.832 & 0.699 & 0.63 & 0.80 \\
w/o stop-gradient & \cellcolor[gray]{0.88}0.923 & \cellcolor[gray]{0.88}0.923 & 0.83 & 0.788 & 0.836 & 0.853 & 0.696 & 0.95 & 0.72 \\
w/o vectors & \cellcolor[gray]{0.88}0.918 & 0.919 & 0.86 & 0.794 & 0.842 & 0.825 & 0.711 & 0.41 & 0.75 \\
w/o USC12K pairs & 0.928 & 0.927 & 0.80 & 0.797 & 0.841 & \cellcolor[gray]{0.88}0.417 & \cellcolor[gray]{0.88}0.292 & 0.75 & 0.74 \\
\midrule
+ joint presence & 0.929 & 0.926 & 0.84 & 0.793 & 0.841 & 0.796 & 0.691 & \cellcolor[gray]{0.88}0.04 & \cellcolor[gray]{0.88}0.12 \\
\bottomrule
\end{tabular}
\caption{\textbf{Ablations.}
Base: UNVEIL trained without the joint presence module.
Middle rows each remove one part from it, and the last row is UNVEIL under the joint query.
Shaded: the cells each part changes most.
Sal., Cam.: salient and camouflaged sets.
SK-L.: SK-LARGE.
Conventions: own-mask IoU on USC12K scenes with both concepts.
Rejection: the absent concept accepted on USC12K scenes with one concept.}
\label{tab:ablation}
\end{table}

\paragraph{Two conventions.}
Because single-concept benchmarks label each concept alone, a single query accepts ``camouflaged object'' on 0.79 of DUTS-TE images.
USC12K gives each region to one concept, and trained on both conventions with plain prompts, ``salient object'' learns to skip animals.
The joint query follows USC12K's convention, and on its scenes with both concepts the two outputs overlap by 0.009 against 0.365 for two single queries.
In these scenes, the camouflaged output of the joint query reaches an IoU of 0.691 with its own mask, against 0.553 for the single query.

\subsection{Ablations}
\label{sec:exp-ablation}

Each middle row of \cref{tab:ablation} removes one part from the base model.
Without the LoRA branch, the BER on ISTD rises from 0.77 to 1.23 and the ODS on SK-LARGE falls from 0.789 to 0.760.
Without replay, the IoU on held-out LVIS categories falls from 0.839 to 0.728, below 0.831 for frozen SAM~3.
Without the stop-gradient, 8 of the 15 test sets score lower.
The mean S-measure on salient sets falls from 0.929 to 0.918 without the vectors, and the own-mask IoUs fall to 0.417 and 0.292 without the USC12K pairs.
Joint presence lowers the wrong acceptance to 0.04 and 0.12.

\section{Conclusion}
\label{sec:conclusion}

SAM~3 sees more than it detects.
On concepts of context and structure, such as camouflaged objects and edges, its semantic path locates the concept while its presence score suppresses the detections.
Promptable concept maps (PCM) ask for this evidence directly, with one phrase in and one map or ``none'' out.
PCM-Bench scores them on concepts of four kinds.
UNVEIL reads the maps from the semantic path and learns presence apart from them.
With one set of weights, it maps regions, thin structures and categories and has the highest score on each task of the PCM-Bench scorecard.
Where a concept lies and whether it is present are thus two questions that a promptable segmenter can answer apart.
Our concepts beyond categories come from seven tasks, and ``none'' is verified for the Context prompts and LVIS categories.
Broader concept vocabularies with verified negatives are the next step.

\section*{Acknowledgments}
This research is supported by the National Research Foundation, Singapore, under its NRF Postdoctoral Award (Project No.~H26-KRO0128).

{
\small
\bibliographystyle{ieeenat_fullname}
\bibliography{refs}

\begin{thebibliography}{136}
\providecommand{\natexlab}[1]{#1}
\providecommand{\url}[1]{\texttt{#1}}
\expandafter\ifx\csname urlstyle\endcsname\relax
  \providecommand{\doi}[1]{doi: #1}\else
  \providecommand{\doi}{doi: \begingroup \urlstyle{rm}\Url}\fi

\bibitem[Alghamdi et~al.(2025)Alghamdi, Usman, Anwar, Bais, and Anwar]{alghamdi2025msrnet}
Leena Alghamdi, Muhammad Usman, Hafeez Anwar, Abdul Bais, and Saeed Anwar.
\newblock {MSRNet}: A multi-scale recursive network for camouflaged object detection.
\newblock \emph{arXiv preprint arXiv:2511.12810}, 2025.

\bibitem[Arbel{\'a}ez et~al.(2011)Arbel{\'a}ez, Maire, Fowlkes, and Malik]{arbelaez2011contour}
Pablo Arbel{\'a}ez, Michael Maire, Charless Fowlkes, and Jitendra Malik.
\newblock Contour detection and hierarchical image segmentation.
\newblock \emph{IEEE Transactions on Pattern Analysis and Machine Intelligence}, 33\penalty0 (5):\penalty0 898--916, 2011.

\bibitem[Bai et~al.(2025)Bai, Chen, Liu, Wang, Ge, Song, Dang, Wang, Wang, Tang, Zhong, Zhu, Yang, Li, Wan, Wang, Ding, Fu, Xu, Ye, Zhang, Xie, Cheng, Zhang, Yang, Xu, and Lin]{bai2025qwen25vl}
Shuai Bai, Keqin Chen, Xuejing Liu, Jialin Wang, Wenbin Ge, Sibo Song, Kai Dang, Peng Wang, Shijie Wang, Jun Tang, Humen Zhong, Yuanzhi Zhu, Mingkun Yang, Zhaohai Li, Jianqiang Wan, Pengfei Wang, Wei Ding, Zheren Fu, Yiheng Xu, Jiabo Ye, Xi Zhang, Tianbao Xie, Zesen Cheng, Hang Zhang, Zhibo Yang, Haiyang Xu, and Junyang Lin.
\newblock {Qwen2.5-VL} technical report.
\newblock \emph{arXiv preprint arXiv:2502.13923}, 2025.

\bibitem[Bolya et~al.(2025)Bolya, Huang, Sun, Cho, Madotto, Wei, Ma, Zhi, Rajasegaran, Bangalath, Wang, Monteiro, Xu, Dong, Ravi, Li, Doll{\'a}r, and Feichtenhofer]{bolya2025perception}
Daniel Bolya, Po-Yao Huang, Peize Sun, Jang~Hyun Cho, Andrea Madotto, Chen Wei, Tengyu Ma, Jiale Zhi, Jathushan Rajasegaran, Hanoona Bangalath, Junke Wang, Marco Monteiro, Hu Xu, Shiyu Dong, Nikhila Ravi, Shang-Wen Li, Piotr Doll{\'a}r, and Christoph Feichtenhofer.
\newblock {Perception Encoder}: The best visual embeddings are not at the output of the network.
\newblock In \emph{Advances in Neural Information Processing Systems (NeurIPS)}, 2025.

\bibitem[Carion et~al.(2020)Carion, Massa, Synnaeve, Usunier, Kirillov, and Zagoruyko]{carion2020end}
Nicolas Carion, Francisco Massa, Gabriel Synnaeve, Nicolas Usunier, Alexander Kirillov, and Sergey Zagoruyko.
\newblock End-to-end object detection with transformers.
\newblock In \emph{Proceedings of the European Conference on Computer Vision (ECCV)}, 2020.

\bibitem[Carion et~al.(2026)Carion, Gustafson, Hu, Debnath, Hu, Suris Coll-Vinent, Ryali, Alwala, Khedr, Huang, Lei, Ma, Guo, Kalla, Marks, Greer, Wang, Sun, R{\"a}dle, Afouras, Mavroudi, Xu, Wu, Zhou, Momeni, Hazra, Ding, Vaze, Porcher, Li, Li, Kamath, Cheng, Doll{\'a}r, Ravi, Saenko, Zhang, and Feichtenhofer]{carion2025sam3}
Nicolas Carion, Laura Gustafson, Yuan-Ting Hu, Shoubhik Debnath, Ronghang Hu, Didac Suris Coll-Vinent, Chaitanya Ryali, Kalyan~Vasudev Alwala, Haitham Khedr, Andrew Huang, Jie Lei, Tengyu Ma, Baishan Guo, Arpit Kalla, Markus Marks, Joseph Greer, Meng Wang, Peize Sun, Roman R{\"a}dle, Triantafyllos Afouras, Effrosyni Mavroudi, Katherine Xu, Tsung-Han Wu, Yu Zhou, Liliane Momeni, Rishi Hazra, Shuangrui Ding, Sagar Vaze, Francois Porcher, Feng Li, Siyuan Li, Aishwarya Kamath, Ho~Kei Cheng, Piotr Doll{\'a}r, Nikhila Ravi, Kate Saenko, Pengchuan Zhang, and Christoph Feichtenhofer.
\newblock {SAM} 3: {Segment Anything} with concepts.
\newblock In \emph{International Conference on Learning Representations (ICLR)}, 2026.

\bibitem[Cetinkaya et~al.(2024)Cetinkaya, Kalkan, and Akbas]{cetinkaya2024ranked}
Bedrettin Cetinkaya, Sinan Kalkan, and Emre Akbas.
\newblock {RankED}: Addressing imbalance and uncertainty in edge detection using ranking-based losses.
\newblock In \emph{Proceedings of the IEEE/CVF Conference on Computer Vision and Pattern Recognition (CVPR)}, pages 3239--3249, 2024.

\bibitem[Chen et~al.(2026{\natexlab{a}})Chen, Zhou, Liu, Jiang, Wu, and Xu]{chen2026liquid}
Ke Chen, Ling Zhou, Yi Liu, Guangqi Jiang, Gengshen Wu, and Shoukun Xu.
\newblock Liquid fusion of heterogeneous representations towards general salient object detection.
\newblock In \emph{Proceedings of the European Conference on Computer Vision (ECCV)}, 2026{\natexlab{a}}.

\bibitem[Chen et~al.(2023)Chen, Zhu, Ding, Cao, Wang, Zhang, Li, Sun, Zang, and Mao]{chen2023sam}
Tianrun Chen, Lanyun Zhu, Chaotao Ding, Runlong Cao, Yan Wang, Shangzhan Zhang, Zejian Li, Lingyun Sun, Ying Zang, and Papa Mao.
\newblock {SAM-Adapter}: Adapting {Segment Anything} in underperformed scenes.
\newblock In \emph{Proceedings of the IEEE/CVF International Conference on Computer Vision (ICCV) Workshops}, pages 3367--3375, 2023.

\bibitem[Chen et~al.(2025{\natexlab{a}})Chen, Cao, Yu, Zhu, Ding, Ji, Chen, Zhu, Xu, Mao, and Zang]{chen2025sam3adapter}
Tianrun Chen, Runlong Cao, Xinda Yu, Lanyun Zhu, Chaotao Ding, Deyi Ji, Cheng Chen, Qi Zhu, Chunyan Xu, Papa Mao, and Ying Zang.
\newblock {SAM3-Adapter}: Efficient adaptation of {Segment Anything} 3 for camouflage object segmentation, shadow detection, and medical image segmentation.
\newblock \emph{arXiv preprint arXiv:2511.19425}, 2025{\natexlab{a}}.

\bibitem[Chen et~al.(2025{\natexlab{b}})Chen, Ren, Dai, Stathaki, and Liu]{chen2025enhancing}
Xuehan Chen, Guangyu Ren, Tianhong Dai, Tania Stathaki, and Hengyan Liu.
\newblock Enhancing prompt generation with adaptive refinement for camouflaged object detection.
\newblock In \emph{Proceedings of the IEEE/CVF International Conference on Computer Vision (ICCV)}, pages 20672--20682, 2025{\natexlab{b}}.

\bibitem[Chen et~al.(2024)Chen, Sun, and Lin]{chen2024camodiffusion}
Zhongxi Chen, Ke Sun, and Xianming Lin.
\newblock {CamoDiffusion}: Camouflaged object detection via conditional diffusion models.
\newblock In \emph{Proceedings of the AAAI Conference on Artificial Intelligence (AAAI)}, pages 1272--1280, 2024.

\bibitem[Chen et~al.(2026{\natexlab{b}})Chen, Zheng, and Ji]{chen2026toward}
Zhigang Chen, Xiawu Zheng, and Rongrong Ji.
\newblock Toward robust in-context segmentation via concept guidance.
\newblock In \emph{Proceedings of the European Conference on Computer Vision (ECCV)}, 2026{\natexlab{b}}.

\bibitem[Cheng et~al.(2026)Cheng, Wu, and Zhou]{cheng2026memo}
Jiaxin Cheng, Yue Wu, and Yicong Zhou.
\newblock {MEMO}: Human-like crisp edge detection using masked edge prediction.
\newblock In \emph{Proceedings of the IEEE/CVF Conference on Computer Vision and Pattern Recognition (CVPR)}, pages 27740--27749, 2026.

\bibitem[Fan et~al.(2017)Fan, Cheng, Liu, Li, and Borji]{fan2017structure}
Deng-Ping Fan, Ming-Ming Cheng, Yun Liu, Tao Li, and Ali Borji.
\newblock Structure-measure: A new way to evaluate foreground maps.
\newblock In \emph{Proceedings of the IEEE International Conference on Computer Vision (ICCV)}, 2017.

\bibitem[Fan et~al.(2020)Fan, Ji, Sun, Cheng, Shen, and Shao]{fan2020camouflaged}
Deng-Ping Fan, Ge-Peng Ji, Guolei Sun, Ming-Ming Cheng, Jianbing Shen, and Ling Shao.
\newblock Camouflaged object detection.
\newblock In \emph{Proceedings of the IEEE/CVF Conference on Computer Vision and Pattern Recognition (CVPR)}, 2020.

\bibitem[Fan et~al.(2022)Fan, Ji, Cheng, and Shao]{fan2022concealed}
Deng-Ping Fan, Ge-Peng Ji, Ming-Ming Cheng, and Ling Shao.
\newblock Concealed object detection.
\newblock \emph{IEEE Transactions on Pattern Analysis and Machine Intelligence}, 44\penalty0 (10):\penalty0 6024--6042, 2022.

\bibitem[Fu et~al.(2026)Fu, Liu, Zhou, Meng, Yang, and He]{fu2026codiff}
Xiaotong Fu, Qian Liu, Qihang Zhou, Wenchao Meng, Qinmin Yang, and Shibo He.
\newblock {CODiff}: One-step diffusion model for camouflaged object detection.
\newblock In \emph{Proceedings of the International Conference on Machine Learning (ICML)}, pages 31977--31991, 2026.

\bibitem[Gao et~al.(2024)Gao, Zhang, Yan, and Lu]{gao2024multi}
Shixuan Gao, Pingping Zhang, Tianyu Yan, and Huchuan Lu.
\newblock Multi-scale and detail-enhanced {Segment Anything Model} for salient object detection.
\newblock In \emph{Proceedings of the ACM International Conference on Multimedia (ACM MM)}, pages 9894--9903, 2024.

\bibitem[Gao et~al.(2026)Gao, Li, He, Li, Zhu, Cheng, and Liu]{gao2026amped}
Yuhan Gao, Xinqing Li, Xin He, Bing Li, Xinzhong Zhu, Ming-Ming Cheng, and Yun Liu.
\newblock {Amped}: Adaptive multi-stage non-edge pruning for edge detection.
\newblock \emph{arXiv preprint arXiv:2603.27661}, 2026.

\bibitem[Gupta et~al.(2019)Gupta, Doll{\'a}r, and Girshick]{gupta2019lvis}
Agrim Gupta, Piotr Doll{\'a}r, and Ross Girshick.
\newblock {LVIS}: A dataset for large vocabulary instance segmentation.
\newblock In \emph{Proceedings of the IEEE/CVF Conference on Computer Vision and Pattern Recognition (CVPR)}, 2019.

\bibitem[Han et~al.(2026{\natexlab{a}})Han, Wu, Sun, and Tang]{han2026beyond}
Jinyu Han, Changguang Wu, Fuming Sun, and Jinhui Tang.
\newblock Beyond appearance: Camouflaged object detection via geometric structure.
\newblock In \emph{Proceedings of the IEEE/CVF Conference on Computer Vision and Pattern Recognition (CVPR)}, pages 25830--25840, 2026{\natexlab{a}}.

\bibitem[Han et~al.(2026{\natexlab{b}})Han, Wu, Sun, Wang, and Tang]{han2026homo}
Jinyu Han, Changguang Wu, Fuming Sun, Mengyin Wang, and Jinhui Tang.
\newblock {H$^2$Net}: Homo- and heterogeneous networks for unified segmentation.
\newblock In \emph{Proceedings of the International Joint Conference on Artificial Intelligence (IJCAI)}, pages 1133--1142, 2026{\natexlab{b}}.

\bibitem[Hao et~al.(2025)Hao, Yu, Liu, Xu, Yue, and Yang]{hao2025simple}
Chao Hao, Zitong Yu, Xin Liu, Jun Xu, Huanjing Yue, and Jingyu Yang.
\newblock A simple yet effective network based on vision transformer for camouflaged object and salient object detection.
\newblock \emph{IEEE Transactions on Image Processing}, 34:\penalty0 608--622, 2025.

\bibitem[He et~al.(2025{\natexlab{a}})He, Zhang, Xiao, Fang, Tang, Zhang, Kong, Fan, Li, and Farsiu]{he2025run}
Chunming He, Rihan Zhang, Fengyang Xiao, Chengyu Fang, Longxiang Tang, Yulun Zhang, Linghe Kong, Deng-Ping Fan, Kai Li, and Sina Farsiu.
\newblock {RUN}: Reversible unfolding network for concealed object segmentation.
\newblock In \emph{Proceedings of the International Conference on Machine Learning (ICML)}, pages 22853--22864, 2025{\natexlab{a}}.

\bibitem[He et~al.(2026{\natexlab{a}})He, Zhang, Tang, Zhang, Zhang, Xiao, Feng, and Farsiu]{he2026nested}
Chunming He, Rihan Zhang, Longxiang Tang, Dingming Zhang, Bojian Zhang, Fengyang Xiao, Jingjia Feng, and Sina Farsiu.
\newblock Nested unfolding network for real-world concealed object segmentation.
\newblock In \emph{Proceedings of the European Conference on Computer Vision (ECCV)}, pages 207--226, 2026{\natexlab{a}}.

\bibitem[He et~al.(2026{\natexlab{b}})He, Zhang, Xiao, Zhang, Cao, and Farsiu]{he2026refining}
Chunming He, Rihan Zhang, Fengyang Xiao, Dingming Zhang, Zhiwen Cao, and Sina Farsiu.
\newblock Refining context-entangled content segmentation via curriculum selection and anti-curriculum promotion.
\newblock In \emph{Proceedings of the International Conference on Machine Learning (ICML)}, pages 42267--42279, 2026{\natexlab{b}}.

\bibitem[He et~al.(2019)He, Zhang, Yang, Shan, and Huang]{he2019bidirectional}
Jianzhong He, Shiliang Zhang, Ming Yang, Yanhu Shan, and Tiejun Huang.
\newblock Bi-directional cascade network for perceptual edge detection.
\newblock In \emph{Proceedings of the IEEE/CVF Conference on Computer Vision and Pattern Recognition (CVPR)}, pages 3828--3837, 2019.

\bibitem[He et~al.(2025{\natexlab{b}})He, Fu, Liu, and Zhao]{he2025samba}
Jiahao He, Keren Fu, Xiaohong Liu, and Qijun Zhao.
\newblock {Samba}: A unified {Mamba}-based framework for general salient object detection.
\newblock In \emph{Proceedings of the IEEE/CVF Conference on Computer Vision and Pattern Recognition (CVPR)}, pages 25314--25324, 2025{\natexlab{b}}.

\bibitem[He et~al.(2023)He, Lin, and Lau]{he2023efficient}
Ruozhen He, Jiaying Lin, and Rynson W.~H. Lau.
\newblock Efficient mirror detection via multi-level heterogeneous learning.
\newblock In \emph{Proceedings of the AAAI Conference on Artificial Intelligence (AAAI)}, pages 790--798, 2023.

\bibitem[Hu et~al.(2022)Hu, Shen, Wallis, Allen-Zhu, Li, Wang, Wang, and Chen]{hu2022lora}
Edward~J. Hu, Yelong Shen, Phillip Wallis, Zeyuan Allen-Zhu, Yuanzhi Li, Shean Wang, Lu Wang, and Weizhu Chen.
\newblock {LoRA}: Low-rank adaptation of large language models.
\newblock In \emph{International Conference on Learning Representations (ICLR)}, 2022.

\bibitem[Huang et~al.(2023{\natexlab{a}})Huang, Dong, Lin, Liu, Lau, and Zuo]{huang2023symmetry}
Tianyu Huang, Bowen Dong, Jiaying Lin, Xiaohui Liu, Rynson W.~H. Lau, and Wangmeng Zuo.
\newblock Symmetry-aware transformer-based mirror detection.
\newblock In \emph{Proceedings of the AAAI Conference on Artificial Intelligence (AAAI)}, pages 935--943, 2023{\natexlab{a}}.

\bibitem[Huang et~al.(2023{\natexlab{b}})Huang, Dai, Xiang, Wang, Chen, Qin, and Xiong]{huang2023feature}
Zhou Huang, Hang Dai, Tian-Zhu Xiang, Shuo Wang, Huai-Xin Chen, Jie Qin, and Huan Xiong.
\newblock Feature shrinkage pyramid for camouflaged object detection with transformers.
\newblock In \emph{Proceedings of the IEEE/CVF Conference on Computer Vision and Pattern Recognition (CVPR)}, pages 5557--5566, 2023{\natexlab{b}}.

\bibitem[Jan et~al.(2025)Jan, El-Maleh, Siddiqui, Bais, and Anwar]{jan2025c3net}
Baber Jan, Aiman~H. El-Maleh, Abdul~Jabbar Siddiqui, Abdul Bais, and Saeed Anwar.
\newblock {C3Net}: Context-contrast network for camouflaged object detection.
\newblock \emph{arXiv preprint arXiv:2511.12627}, 2025.

\bibitem[Ji et~al.(2023)Ji, Fan, Xu, Cheng, Zhou, and Van~Gool]{ji2023sam}
Ge-Peng Ji, Deng-Ping Fan, Peng Xu, Ming-Ming Cheng, Bowen Zhou, and Luc Van~Gool.
\newblock {SAM} struggles in concealed scenes -- empirical study on ``{Segment Anything}''.
\newblock \emph{Science China Information Sciences}, 66:\penalty0 226101, 2023.

\bibitem[Jie et~al.(2024)Jie, Guo, Wu, Wu, and Hua]{jie2024edgenat}
Jinghuai Jie, Yan Guo, Guixing Wu, Junmin Wu, and Baojian Hua.
\newblock {EdgeNAT}: Transformer for efficient edge detection.
\newblock In \emph{Proceedings of the European Conference on Artificial Intelligence (ECAI)}, pages 435--442, 2024.

\bibitem[Jie and Zhang(2025)]{jie2025shadowadapter}
Leiping Jie and Hui Zhang.
\newblock {ShadowAdapter}: Adapting {Segment Anything Model} with auto-prompt for shadow detection.
\newblock \emph{Expert Systems with Applications}, 273:\penalty0 126809, 2025.

\bibitem[Ke et~al.(2017)Ke, Chen, Jiao, Zhao, and Ye]{ke2017srn}
Wei Ke, Jie Chen, Jianbin Jiao, Guoying Zhao, and Qixiang Ye.
\newblock {SRN}: Side-output residual network for object symmetry detection in the wild.
\newblock In \emph{Proceedings of the IEEE Conference on Computer Vision and Pattern Recognition (CVPR)}, 2017.

\bibitem[Kim et~al.(2022)Kim, Kim, Lee, Cha, Lee, and Kim]{kim2022revisiting}
Taehun Kim, Kunhee Kim, Joonyeong Lee, Dongmin Cha, Jiho Lee, and Daijin Kim.
\newblock Revisiting image pyramid structure for high resolution salient object detection.
\newblock In \emph{Proceedings of the Asian Conference on Computer Vision (ACCV)}, pages 257--273, 2022.

\bibitem[Kirillov et~al.(2023)Kirillov, Mintun, Ravi, Mao, Rolland, Gustafson, Xiao, Whitehead, Berg, Lo, Doll{\'a}r, and Girshick]{kirillov2023segment}
Alexander Kirillov, Eric Mintun, Nikhila Ravi, Hanzi Mao, Chloe Rolland, Laura Gustafson, Tete Xiao, Spencer Whitehead, Alexander~C. Berg, Wan-Yen Lo, Piotr Doll{\'a}r, and Ross Girshick.
\newblock {Segment Anything}.
\newblock In \emph{Proceedings of the IEEE/CVF International Conference on Computer Vision (ICCV)}, pages 4015--4026, 2023.

\bibitem[Kumar et~al.(2022)Kumar, Raghunathan, Jones, Ma, and Liang]{kumar2022fine}
Ananya Kumar, Aditi Raghunathan, Robbie Jones, Tengyu Ma, and Percy Liang.
\newblock Fine-tuning can distort pretrained features and underperform out-of-distribution.
\newblock In \emph{International Conference on Learning Representations (ICLR)}, 2022.

\bibitem[Lai et~al.(2024)Lai, Tian, Chen, Li, Yuan, Liu, and Jia]{lai2024lisa}
Xin Lai, Zhuotao Tian, Yukang Chen, Yanwei Li, Yuhui Yuan, Shu Liu, and Jiaya Jia.
\newblock {LISA}: Reasoning segmentation via large language model.
\newblock In \emph{Proceedings of the IEEE/CVF Conference on Computer Vision and Pattern Recognition (CVPR)}, pages 9579--9589, 2024.

\bibitem[Le et~al.(2019)Le, Nguyen, Nie, Tran, and Sugimoto]{le2019anabranch}
Trung-Nghia Le, Tam~V. Nguyen, Zhongliang Nie, Minh-Triet Tran, and Akihiro Sugimoto.
\newblock Anabranch network for camouflaged object segmentation.
\newblock \emph{Computer Vision and Image Understanding}, 184:\penalty0 45--56, 2019.

\bibitem[Li and Yu(2015)]{li2015visual}
Guanbin Li and Yizhou Yu.
\newblock Visual saliency based on multiscale deep features.
\newblock In \emph{Proceedings of the IEEE Conference on Computer Vision and Pattern Recognition (CVPR)}, 2015.

\bibitem[Li et~al.(2025{\natexlab{a}})Li, Zhang, Wang, Deng, Wang, Meng, and Cao]{li2025segearth}
Kaiyu Li, Shengqi Zhang, Yujie Wang, Yupeng Deng, Zhi Wang, Deyu Meng, and Xiangyong Cao.
\newblock {SegEarth-OV3}: Exploring {SAM} 3 for open-vocabulary semantic segmentation in remote sensing images.
\newblock \emph{arXiv preprint arXiv:2512.08730}, 2025{\natexlab{a}}.

\bibitem[Li et~al.(2025{\natexlab{b}})Li, Ye, Xiong, Bai, Tang, Song, Xing, Ju, Li, and Zhu]{li2025glasswizard}
Wenxue Li, Tian Ye, Xinyu Xiong, Jinbin Bai, Feilong Tang, Wenxuan Song, Zhaohu Xing, Lie Ju, Guanbin Li, and Lei Zhu.
\newblock {GlassWizard}: Harvesting diffusion priors for glass surface detection.
\newblock In \emph{Proceedings of the IEEE/CVF International Conference on Computer Vision (ICCV)}, pages 17848--17858, 2025{\natexlab{b}}.

\bibitem[Li et~al.(2014)Li, Hou, Koch, Rehg, and Yuille]{li2014secrets}
Yin Li, Xiaodi Hou, Christof Koch, James~M. Rehg, and Alan~L. Yuille.
\newblock The secrets of salient object segmentation.
\newblock In \emph{Proceedings of the IEEE Conference on Computer Vision and Pattern Recognition (CVPR)}, 2014.

\bibitem[Li et~al.(2025{\natexlab{c}})Li, Poma, Bai, Xiao, Yang, Li, and Li]{li2025edmb}
Yachuan Li, Xavier~Soria Poma, Yun Bai, Qian Xiao, Chaozhi Yang, Guanlin Li, and Zongmin Li.
\newblock {EDMB}: Edge detector with {Mamba}.
\newblock In \emph{Proceedings of the IEEE/CVF Winter Conference on Applications of Computer Vision (WACV)}, pages 7671--7680, 2025{\natexlab{c}}.

\bibitem[Li et~al.(2026{\natexlab{a}})Li, Poma, Xi, Li, Yang, Xiao, Bai, and Li]{li2026new}
Yachuan Li, Xavier~Soria Poma, Yongke Xi, Guanlin Li, Chaozhi Yang, Qian Xiao, Yun Bai, and Zongmin Li.
\newblock A new baseline for edge detection: Make encoder--decoder great again.
\newblock \emph{Signal Processing: Image Communication}, 142:\penalty0 117485, 2026{\natexlab{a}}.

\bibitem[Li et~al.(2026{\natexlab{b}})Li, Zhang, Zhao, Liu, and Li]{li2026dganet}
Yuetong Li, Qing Zhang, Yilin Zhao, Zeming Liu, and Gongyang Li.
\newblock {DGA-Net}: Enhancing {SAM} with depth prompting and graph-anchor guidance for camouflaged object detection.
\newblock \emph{IEEE Transactions on Circuits and Systems for Video Technology}, 2026{\natexlab{b}}.

\bibitem[Li and Hoiem(2018)]{li2018learning}
Zhizhong Li and Derek Hoiem.
\newblock Learning without forgetting.
\newblock \emph{IEEE Transactions on Pattern Analysis and Machine Intelligence}, 40\penalty0 (12):\penalty0 2935--2947, 2018.

\bibitem[Liang et~al.(2026)Liang, Wang, Li, Liu, and Wang]{liang2026pixels}
Shuang Liang, Zeqing Wang, Yuxian Li, Xihui Liu, and Han Wang.
\newblock From pixels to concepts: Do segmentation models understand what they segment?
\newblock In \emph{Advances in Neural Information Processing Systems (NeurIPS)}, 2026.

\bibitem[Liao et~al.(2026)Liao, Zhu, Ma, Jiang, and Dai]{liao2026efficient}
Wenbin Liao, Hao Zhu, Yike Ma, Hao Jiang, and Feng Dai.
\newblock Efficient {SAM} 3 adaptation for multi-class semantic segmentation via dense competitive representations.
\newblock In \emph{Advances in Neural Information Processing Systems (NeurIPS)}, 2026.

\bibitem[Liu et~al.(2021)Liu, Tian, Chen, Jiao, and Ye]{liu2021adaptive}
Chang Liu, Yunjie Tian, Zhiwen Chen, Jianbin Jiao, and Qixiang Ye.
\newblock Adaptive linear span network for object skeleton detection.
\newblock \emph{IEEE Transactions on Image Processing}, 30:\penalty0 5096--5108, 2021.

\bibitem[Liu et~al.(2023{\natexlab{a}})Liu, Ding, and Jiang]{liu2023gres}
Chang Liu, Henghui Ding, and Xudong Jiang.
\newblock {GRES}: Generalized referring expression segmentation.
\newblock In \emph{Proceedings of the IEEE/CVF Conference on Computer Vision and Pattern Recognition (CVPR)}, pages 23592--23601, 2023{\natexlab{a}}.

\bibitem[Liu et~al.(2020)Liu, Hou, and Cheng]{liu2020dynamic}
Jiang-Jiang Liu, Qibin Hou, and Ming-Ming Cheng.
\newblock Dynamic feature integration for simultaneous detection of salient object, edge and skeleton.
\newblock \emph{IEEE Transactions on Image Processing}, 29:\penalty0 8652--8667, 2020.

\bibitem[Liu et~al.(2024{\natexlab{a}})Liu, Luo, Zhang, and Han]{liu2024vst}
Nian Liu, Ziyang Luo, Ni Zhang, and Junwei Han.
\newblock {VST++}: Efficient and stronger visual saliency transformer.
\newblock \emph{IEEE Transactions on Pattern Analysis and Machine Intelligence}, 46\penalty0 (11):\penalty0 7300--7316, 2024{\natexlab{a}}.

\bibitem[Liu et~al.(2024{\natexlab{b}})Liu, Zeng, Ren, Li, Zhang, Yang, Jiang, Li, Yang, Su, Zhu, and Zhang]{liu2024grounding}
Shilong Liu, Zhaoyang Zeng, Tianhe Ren, Feng Li, Hao Zhang, Jie Yang, Qing Jiang, Chunyuan Li, Jianwei Yang, Hang Su, Jun Zhu, and Lei Zhang.
\newblock {Grounding DINO}: Marrying {DINO} with grounded pre-training for open-set object detection.
\newblock In \emph{Proceedings of the European Conference on Computer Vision (ECCV)}, 2024{\natexlab{b}}.

\bibitem[Liu et~al.(2023{\natexlab{b}})Liu, Shen, Pun, and Cun]{liu2023explicit}
Weihuang Liu, Xi Shen, Chi-Man Pun, and Xiaodong Cun.
\newblock Explicit visual prompting for low-level structure segmentations.
\newblock In \emph{Proceedings of the IEEE/CVF Conference on Computer Vision and Pattern Recognition (CVPR)}, pages 19434--19445, 2023{\natexlab{b}}.

\bibitem[Liu et~al.(2025{\natexlab{a}})Liu, Wang, and Gao]{liu2025first}
Wutao Liu, YiDan Wang, and Pan Gao.
\newblock First {RAG}, second {SEG}: A training-free paradigm for camouflaged object detection.
\newblock \emph{arXiv preprint arXiv:2508.15313}, 2025{\natexlab{a}}.

\bibitem[Liu et~al.(2026)Liu, Shen, Pun, and Cun]{liu2026explicit}
Weihuang Liu, Xi Shen, Chi-Man Pun, and Xiaodong Cun.
\newblock Explicit visual prompting for universal foreground segmentations.
\newblock \emph{IEEE Transactions on Pattern Analysis and Machine Intelligence}, 48\penalty0 (2):\penalty0 1762--1777, 2026.

\bibitem[Liu et~al.(2017)Liu, Cheng, Hu, Wang, and Bai]{liu2017richer}
Yun Liu, Ming-Ming Cheng, Xiaowei Hu, Kai Wang, and Xiang Bai.
\newblock Richer convolutional features for edge detection.
\newblock In \emph{Proceedings of the IEEE Conference on Computer Vision and Pattern Recognition (CVPR)}, 2017.

\bibitem[Liu et~al.(2024{\natexlab{c}})Liu, Zhang, Wang, Wang, Yang, and Tang]{liu2024universal}
Yong Liu, Cairong Zhang, Yitong Wang, Jiahao Wang, Yujiu Yang, and Yansong Tang.
\newblock Universal segmentation at arbitrary granularity with language instruction.
\newblock In \emph{Proceedings of the IEEE/CVF Conference on Computer Vision and Pattern Recognition (CVPR)}, pages 3459--3469, 2024{\natexlab{c}}.

\bibitem[Liu et~al.(2025{\natexlab{b}})Liu, Li, Dong, Li, Zhang, Xu, and Han]{liu2025seamless}
Yi Liu, Chengxin Li, Xiaohui Dong, Lei Li, Dingwen Zhang, Shoukun Xu, and Jungong Han.
\newblock Seamless detection: Unifying salient object detection and camouflaged object detection.
\newblock \emph{Expert Systems with Applications}, 274:\penalty0 126912, 2025{\natexlab{b}}.

\bibitem[Liufu et~al.(2025)Liufu, Tan, Lin, Qi, Li, and Hu]{liufu2025sauge}
Xing Liufu, Chaolei Tan, Xiaotong Lin, Yonggang Qi, Jinxuan Li, and Jian-Fang Hu.
\newblock {SAUGE}: Taming {SAM} for uncertainty-aligned multi-granularity edge detection.
\newblock In \emph{Proceedings of the AAAI Conference on Artificial Intelligence (AAAI)}, pages 5766--5774, 2025.

\bibitem[L{\"u}ddecke and Ecker(2022)]{luddecke2022image}
Timo L{\"u}ddecke and Alexander Ecker.
\newblock Image segmentation using text and image prompts.
\newblock In \emph{Proceedings of the IEEE/CVF Conference on Computer Vision and Pattern Recognition (CVPR)}, pages 7086--7096, 2022.

\bibitem[Luo et~al.(2024)Luo, Liu, Zhao, Yang, Zhang, Fan, Khan, and Han]{luo2024vscode}
Ziyang Luo, Nian Liu, Wangbo Zhao, Xuguang Yang, Dingwen Zhang, Deng-Ping Fan, Fahad Khan, and Junwei Han.
\newblock {VSCode}: General visual salient and camouflaged object detection with {2D} prompt learning.
\newblock In \emph{Proceedings of the IEEE/CVF Conference on Computer Vision and Pattern Recognition (CVPR)}, pages 17169--17180, 2024.

\bibitem[Lv et~al.(2021)Lv, Zhang, Dai, Li, Liu, Barnes, and Fan]{lv2021simultaneously}
Yunqiu Lv, Jing Zhang, Yuchao Dai, Aixuan Li, Bowen Liu, Nick Barnes, and Deng-Ping Fan.
\newblock Simultaneously localize, segment and rank the camouflaged objects.
\newblock In \emph{Proceedings of the IEEE/CVF Conference on Computer Vision and Pattern Recognition (CVPR)}, pages 11591--11601, 2021.

\bibitem[Nakamura et~al.(2026)Nakamura, Iino, Okada, and Taniguchi]{nakamura2026easycontroledge}
Hiroki Nakamura, Hiroto Iino, Masashi Okada, and Tadahiro Taniguchi.
\newblock {EasyControlEdge}: A foundation-model fine-tuning for edge detection.
\newblock \emph{arXiv preprint arXiv:2602.16238}, 2026.

\bibitem[Pang et~al.(2024)Pang, Zhao, Xiang, Zhang, and Lu]{pang2024zoomnext}
Youwei Pang, Xiaoqi Zhao, Tian-Zhu Xiang, Lihe Zhang, and Huchuan Lu.
\newblock {ZoomNeXt}: A unified collaborative pyramid network for camouflaged object detection.
\newblock \emph{IEEE Transactions on Pattern Analysis and Machine Intelligence}, 46\penalty0 (12):\penalty0 9205--9220, 2024.

\bibitem[Prati et~al.(2003)Prati, Miki{\'c}, Trivedi, and Cucchiara]{prati2003detecting}
Andrea Prati, Ivana Miki{\'c}, Mohan~M. Trivedi, and Rita Cucchiara.
\newblock Detecting moving shadows: Algorithms and evaluation.
\newblock \emph{IEEE Transactions on Pattern Analysis and Machine Intelligence}, 25\penalty0 (7):\penalty0 918--923, 2003.

\bibitem[Pu et~al.(2022)Pu, Huang, Liu, Guan, and Ling]{pu2022edter}
Mengyang Pu, Yaping Huang, Yuming Liu, Qingji Guan, and Haibin Ling.
\newblock {EDTER}: Edge detection with transformer.
\newblock In \emph{Proceedings of the IEEE/CVF Conference on Computer Vision and Pattern Recognition (CVPR)}, pages 1402--1412, 2022.

\bibitem[Rasheed et~al.(2024)Rasheed, Maaz, Shaji, Shaker, Khan, Cholakkal, Anwer, Xing, Yang, and Khan]{rasheed2024glamm}
Hanoona Rasheed, Muhammad Maaz, Sahal Shaji, Abdelrahman Shaker, Salman Khan, Hisham Cholakkal, Rao~M. Anwer, Eric Xing, Ming-Hsuan Yang, and Fahad~S. Khan.
\newblock {GLaMM}: Pixel grounding large multimodal model.
\newblock In \emph{Proceedings of the IEEE/CVF Conference on Computer Vision and Pattern Recognition (CVPR)}, pages 13009--13018, 2024.

\bibitem[Ravi et~al.(2025)Ravi, Gabeur, Hu, Hu, Ryali, Ma, Khedr, R{\"a}dle, Rolland, Gustafson, Mintun, Pan, Alwala, Carion, Wu, Girshick, Doll{\'a}r, and Feichtenhofer]{ravi2025sam2}
Nikhila Ravi, Valentin Gabeur, Yuan-Ting Hu, Ronghang Hu, Chaitanya Ryali, Tengyu Ma, Haitham Khedr, Roman R{\"a}dle, Chloe Rolland, Laura Gustafson, Eric Mintun, Junting Pan, Kalyan~Vasudev Alwala, Nicolas Carion, Chao-Yuan Wu, Ross Girshick, Piotr Doll{\'a}r, and Christoph Feichtenhofer.
\newblock {SAM} 2: {Segment Anything} in images and videos.
\newblock In \emph{International Conference on Learning Representations (ICLR)}, 2025.

\bibitem[Rei{\ss} et~al.(2025)Rei{\ss}, Marinov, Jaus, Seibold, Sarfraz, Rodner, and Stiefelhagen]{reiss2025visual}
Simon Rei{\ss}, Zdravko Marinov, Alexander Jaus, Constantin Seibold, M.~Saquib Sarfraz, Erik Rodner, and Rainer Stiefelhagen.
\newblock Is visual in-context learning for compositional medical tasks within reach?
\newblock In \emph{Proceedings of the IEEE/CVF International Conference on Computer Vision (ICCV)}, pages 2642--2652, 2025.

\bibitem[Ren et~al.(2024{\natexlab{a}})Ren, Liu, Zeng, Lin, Li, Cao, Chen, Huang, Chen, Yan, Zeng, Zhang, Li, Yang, Li, Jiang, and Zhang]{ren2024grounded}
Tianhe Ren, Shilong Liu, Ailing Zeng, Jing Lin, Kunchang Li, He Cao, Jiayu Chen, Xinyu Huang, Yukang Chen, Feng Yan, Zhaoyang Zeng, Hao Zhang, Feng Li, Jie Yang, Hongyang Li, Qing Jiang, and Lei Zhang.
\newblock Grounded {SAM}: Assembling open-world models for diverse visual tasks.
\newblock \emph{arXiv preprint arXiv:2401.14159}, 2024{\natexlab{a}}.

\bibitem[Ren et~al.(2024{\natexlab{b}})Ren, Huang, Wei, Zhao, Fu, Feng, and Jin]{ren2024pixellm}
Zhongwei Ren, Zhicheng Huang, Yunchao Wei, Yao Zhao, Dongmei Fu, Jiashi Feng, and Xiaojie Jin.
\newblock {PixelLM}: Pixel reasoning with large multimodal model.
\newblock In \emph{Proceedings of the IEEE/CVF Conference on Computer Vision and Pattern Recognition (CVPR)}, pages 26374--26383, 2024{\natexlab{b}}.

\bibitem[Rolnick et~al.(2019)Rolnick, Ahuja, Schwarz, Lillicrap, and Wayne]{rolnick2019experience}
David Rolnick, Arun Ahuja, Jonathan Schwarz, Timothy Lillicrap, and Gregory Wayne.
\newblock Experience replay for continual learning.
\newblock In \emph{Advances in Neural Information Processing Systems (NeurIPS)}, 2019.

\bibitem[Rust and Kambhamettu(2026)]{rust2026resolution}
Tyler Rust and Chandra Kambhamettu.
\newblock When does resolution help a frozen backbone? {G}lobal attention at resolution predicts scalable adaptation for camouflaged and marine animal segmentation.
\newblock \emph{arXiv preprint arXiv:2607.02708}, 2026.

\bibitem[Rust et~al.(2026)Rust, McNally, O'Donnell, Kelly, and Kambhamettu]{rust2026bedsam2}
Tyler Rust, Dara McNally, Kyle O'Donnell, Colin Kelly, and Chandra Kambhamettu.
\newblock {BED-SAM2}: Boundary-enhanced-depth {SAM2} via monocular geometric priors.
\newblock \emph{arXiv preprint arXiv:2605.24893}, 2026.

\bibitem[Sajjan et~al.(2020)Sajjan, Moore, Pan, Nagaraja, Lee, Zeng, and Song]{sajjan2020cleargrasp}
Shreeyak~S. Sajjan, Matthew Moore, Mike Pan, Ganesh Nagaraja, Johnny Lee, Andy Zeng, and Shuran Song.
\newblock {ClearGrasp}: {3D} shape estimation of transparent objects for manipulation.
\newblock In \emph{IEEE International Conference on Robotics and Automation (ICRA)}, 2020.

\bibitem[Shao et~al.(2025)Shao, Chen, Shi, Liu, Li, Ma, and Yeung]{shao2025mirror}
Zhiwen Shao, Rui Chen, Xuehuai Shi, Bing Liu, Canlin Li, Lizhuang Ma, and Dit-Yan Yeung.
\newblock Mirror detection via multi-directional similarity perception and spectral saliency enhancement.
\newblock \emph{IEEE Transactions on Circuits and Systems for Video Technology}, 35\penalty0 (10):\penalty0 10099--10109, 2025.

\bibitem[Shen et~al.(2017)Shen, Zhao, Jiang, Wang, Bai, and Yuille]{shen2017deepskeleton}
Wei Shen, Kai Zhao, Yuan Jiang, Yan Wang, Xiang Bai, and Alan Yuille.
\newblock {DeepSkeleton}: Learning multi-task scale-associated deep side outputs for object skeleton extraction in natural images.
\newblock \emph{IEEE Transactions on Image Processing}, 26\penalty0 (11):\penalty0 5298--5311, 2017.

\bibitem[Su et~al.(2021)Su, Liu, Yu, Hu, Liao, Tian, Pietik{\"a}inen, and Liu]{su2021pixel}
Zhuo Su, Wenzhe Liu, Zitong Yu, Dewen Hu, Qing Liao, Qi Tian, Matti Pietik{\"a}inen, and Li Liu.
\newblock Pixel difference networks for efficient edge detection.
\newblock In \emph{Proceedings of the IEEE/CVF International Conference on Computer Vision (ICCV)}, pages 5117--5127, 2021.

\bibitem[Sung et~al.(2022)Sung, Cho, and Bansal]{sung2022lst}
Yi-Lin Sung, Jaemin Cho, and Mohit Bansal.
\newblock {LST}: Ladder side-tuning for parameter and memory efficient transfer learning.
\newblock In \emph{Advances in Neural Information Processing Systems (NeurIPS)}, pages 12991--13005, 2022.

\bibitem[Tan et~al.(2021)Tan, Lin, Chang, and Savva]{tan2021mirror3d}
Jiaqi Tan, Weijie Lin, Angel~X. Chang, and Manolis Savva.
\newblock {Mirror3D}: Depth refinement for mirror surfaces.
\newblock In \emph{Proceedings of the IEEE/CVF Conference on Computer Vision and Pattern Recognition (CVPR)}, pages 15990--15999, 2021.

\bibitem[Tien and Shen(2026)]{tien2026activesam}
Tran~Dinh Tien and Zhiqiang Shen.
\newblock {ActiveSAM}: Fast and accurate open-vocabulary semantic segmentation with frozen {SAM} 3.
\newblock \emph{arXiv preprint arXiv:2606.16996}, 2026.

\bibitem[Vicente et~al.(2016)Vicente, Hou, Yu, Hoai, and Samaras]{vicente2016large}
Tom{\'a}s F.~Yago Vicente, Le Hou, Chen-Ping Yu, Minh Hoai, and Dimitris Samaras.
\newblock Large-scale training of shadow detectors with noisily-annotated shadow examples.
\newblock In \emph{Proceedings of the European Conference on Computer Vision (ECCV)}, pages 816--832, 2016.

\bibitem[Wang et~al.(2018)Wang, Li, and Yang]{wang2018stacked}
Jifeng Wang, Xiang Li, and Jian Yang.
\newblock Stacked conditional generative adversarial networks for jointly learning shadow detection and shadow removal.
\newblock In \emph{Proceedings of the IEEE Conference on Computer Vision and Pattern Recognition (CVPR)}, 2018.

\bibitem[Wang et~al.(2025{\natexlab{a}})Wang, Qin, Lu, Wang, and Tao]{wang2025assisted}
Kuan Wang, Yanjun Qin, Mengge Lu, Liejun Wang, and Xiaoming Tao.
\newblock Assisted refinement network based on channel information interaction for camouflaged and salient object detection.
\newblock \emph{arXiv preprint arXiv:2512.11369}, 2025{\natexlab{a}}.

\bibitem[Wang et~al.(2025{\natexlab{b}})Wang, Tu, Li, Liu, and Luo]{wang2025unified}
Kunpeng Wang, Zhengzheng Tu, Chenglong Li, Zhengyi Liu, and Bin Luo.
\newblock Unified-modal salient object detection via adaptive prompt learning.
\newblock \emph{IEEE Transactions on Circuits and Systems for Video Technology}, 35\penalty0 (12):\penalty0 12109--12124, 2025{\natexlab{b}}.

\bibitem[Wang et~al.(2017)Wang, Lu, Wang, Feng, Wang, Yin, and Ruan]{wang2017learning}
Lijun Wang, Huchuan Lu, Yifan Wang, Mengyang Feng, Dong Wang, Baocai Yin, and Xiang Ruan.
\newblock Learning to detect salient objects with image-level supervision.
\newblock In \emph{Proceedings of the IEEE Conference on Computer Vision and Pattern Recognition (CVPR)}, 2017.

\bibitem[Wang et~al.(2023{\natexlab{a}})Wang, Zhang, Cao, Wang, Shen, and Huang]{wang2023seggpt}
Xinlong Wang, Xiaosong Zhang, Yue Cao, Wen Wang, Chunhua Shen, and Tiejun Huang.
\newblock {SegGPT}: Towards segmenting everything in context.
\newblock In \emph{Proceedings of the IEEE/CVF International Conference on Computer Vision (ICCV)}, pages 1130--1140, 2023{\natexlab{a}}.

\bibitem[Wang et~al.(2019)Wang, Xu, Tsogkas, Bai, Dickinson, and Siddiqi]{wang2019deepflux}
Yukang Wang, Yongchao Xu, Stavros Tsogkas, Xiang Bai, Sven Dickinson, and Kaleem Siddiqi.
\newblock {DeepFlux} for skeletons in the wild.
\newblock In \emph{Proceedings of the IEEE/CVF Conference on Computer Vision and Pattern Recognition (CVPR)}, pages 5287--5296, 2019.

\bibitem[Wang et~al.(2023{\natexlab{b}})Wang, Wang, Fan, Wang, and He]{wang2023pixels}
Yi Wang, Ruili Wang, Xin Fan, Tianzhu Wang, and Xiangjian He.
\newblock Pixels, regions, and objects: Multiple enhancement for salient object detection.
\newblock In \emph{Proceedings of the IEEE/CVF Conference on Computer Vision and Pattern Recognition (CVPR)}, pages 10031--10040, 2023{\natexlab{b}}.

\bibitem[Wang et~al.(2026)Wang, Liu, Zhou, Feng, and Li]{wang2026revisiting}
Yonghui Wang, Shaokai Liu, Wengang Zhou, Hao Feng, and Houqiang Li.
\newblock Revisiting shadow detection from a vision-language perspective.
\newblock \emph{arXiv preprint arXiv:2605.11771}, 2026.

\bibitem[Wei et~al.(2020)Wei, Wang, and Huang]{wei2020f3net}
Jun Wei, Shuhui Wang, and Qingming Huang.
\newblock {F$^3$Net}: Fusion, feedback and focus for salient object detection.
\newblock In \emph{Proceedings of the AAAI Conference on Artificial Intelligence (AAAI)}, pages 12321--12328, 2020.

\bibitem[Wortsman et~al.(2022)Wortsman, Ilharco, Kim, Li, Kornblith, Roelofs, Lopes, Hajishirzi, Farhadi, Namkoong, and Schmidt]{wortsman2022robust}
Mitchell Wortsman, Gabriel Ilharco, Jong~Wook Kim, Mike Li, Simon Kornblith, Rebecca Roelofs, Raphael~Gontijo Lopes, Hannaneh Hajishirzi, Ali Farhadi, Hongseok Namkoong, and Ludwig Schmidt.
\newblock Robust fine-tuning of zero-shot models.
\newblock In \emph{Proceedings of the IEEE/CVF Conference on Computer Vision and Pattern Recognition (CVPR)}, pages 7959--7971, 2022.

\bibitem[Wu et~al.(2024)Wu, Zhong, Xing, Lai, Liu, Chen, Wang, Zhu, Lu, Lu, Luo, Qiao, and Dai]{wu2024visionllm}
Jiannan Wu, Muyan Zhong, Sen Xing, Zeqiang Lai, Zhaoyang Liu, Zhe Chen, Wenhai Wang, Xizhou Zhu, Lewei Lu, Tong Lu, Ping Luo, Yu Qiao, and Jifeng Dai.
\newblock {VisionLLM} v2: An end-to-end generalist multimodal large language model for hundreds of vision-language tasks.
\newblock In \emph{Advances in Neural Information Processing Systems (NeurIPS)}, pages 69925--69975, 2024.

\bibitem[Wu et~al.(2022)Wu, Liu, Zhang, Cheng, and Ren]{wu2022edn}
Yu-Huan Wu, Yun Liu, Le Zhang, Ming-Ming Cheng, and Bo Ren.
\newblock {EDN}: Salient object detection via extremely-downsampled network.
\newblock \emph{IEEE Transactions on Image Processing}, 31:\penalty0 3125--3136, 2022.

\bibitem[Xia et~al.(2024)Xia, Han, Han, Pan, Song, and Huang]{xia2024gsva}
Zhuofan Xia, Dongchen Han, Yizeng Han, Xuran Pan, Shiji Song, and Gao Huang.
\newblock {GSVA}: Generalized segmentation via multimodal large language models.
\newblock In \emph{Proceedings of the IEEE/CVF Conference on Computer Vision and Pattern Recognition (CVPR)}, pages 3858--3869, 2024.

\bibitem[Xie et~al.(2020)Xie, Wang, Wang, Ding, Shen, and Luo]{xie2020segmenting}
Enze Xie, Wenjia Wang, Wenhai Wang, Mingyu Ding, Chunhua Shen, and Ping Luo.
\newblock Segmenting transparent objects in the wild.
\newblock In \emph{Proceedings of the European Conference on Computer Vision (ECCV)}, 2020.

\bibitem[Xie and Tu(2015)]{xie2015holistically}
Saining Xie and Zhuowen Tu.
\newblock Holistically-nested edge detection.
\newblock In \emph{Proceedings of the IEEE International Conference on Computer Vision (ICCV)}, 2015.

\bibitem[Xie et~al.(2024)Xie, Wang, Yu, Tan, and Xie]{xie2024csfwinformer}
Zhifeng Xie, Sen Wang, Qiucheng Yu, Xin Tan, and Yuan Xie.
\newblock {CSFwinformer}: Cross-space-frequency window transformer for mirror detection.
\newblock \emph{IEEE Transactions on Image Processing}, 33:\penalty0 1853--1867, 2024.

\bibitem[Xiong et~al.(2025{\natexlab{a}})Xiong, Wu, Lu, and Xia]{xiong2025sam3unet}
Xinyu Xiong, Zihuang Wu, Lei Lu, and Yufa Xia.
\newblock {SAM3-UNet}: Simplified adaptation of {Segment Anything Model} 3.
\newblock \emph{arXiv preprint arXiv:2512.01789}, 2025{\natexlab{a}}.

\bibitem[Xiong et~al.(2025{\natexlab{b}})Xiong, Wu, Zhang, Lu, Li, and Li]{xiong2025sam2unext}
Xinyu Xiong, Zihuang Wu, Lei Zhang, Lei Lu, Ming Li, and Guanbin Li.
\newblock {SAM2-UNeXT}: An improved high-resolution baseline for adapting foundation models to downstream segmentation tasks.
\newblock \emph{arXiv preprint arXiv:2508.03566}, 2025{\natexlab{b}}.

\bibitem[Xiong et~al.(2026)Xiong, Wu, Tan, Li, Tang, Chen, Li, Ma, and Li]{xiong2026sam2unet}
Xinyu Xiong, Zihuang Wu, Shuangyi Tan, Wenxue Li, Feilong Tang, Ying Chen, Siying Li, Jie Ma, and Guanbin Li.
\newblock {SAM2-UNet}: {Segment Anything} 2 makes strong encoder for natural and medical image segmentation.
\newblock \emph{Visual Intelligence}, 4:\penalty0 2, 2026.

\bibitem[Xu et~al.(2027)Xu, Zhou, Li, Shi, Yi, and Yu]{xu2027hvpnet}
Jiawei Xu, Qiangqiang Zhou, Zhouping Li, Yanjiao Shi, Yugen Yi, and Jiacong Yu.
\newblock {HVPNet}: A bio-inspired network for general salient and camouflaged object detection.
\newblock \emph{Neural Networks}, 205:\penalty0 109340, 2027.

\bibitem[Xu et~al.(2025)Xu, Yang, and Xu]{xu2025dsunet}
Yimin Xu, Fan Yang, and Bin Xu.
\newblock {DSU-Net}: An improved {U-Net} model based on {DINOv2} and {SAM2} with multi-scale cross-model feature enhancement.
\newblock In \emph{Proceedings of the International Conference on Artificial Intelligence and Big Data (ICAIBD)}, pages 625--632, 2025.

\bibitem[Yan et~al.(2013)Yan, Xu, Shi, and Jia]{yan2013hierarchical}
Qiong Yan, Li Xu, Jianping Shi, and Jiaya Jia.
\newblock Hierarchical saliency detection.
\newblock In \emph{Proceedings of the IEEE Conference on Computer Vision and Pattern Recognition (CVPR)}, 2013.

\bibitem[Yang et~al.(2013)Yang, Zhang, Lu, Ruan, and Yang]{yang2013saliency}
Chuan Yang, Lihe Zhang, Huchuan Lu, Xiang Ruan, and Ming-Hsuan Yang.
\newblock Saliency detection via graph-based manifold ranking.
\newblock In \emph{Proceedings of the IEEE Conference on Computer Vision and Pattern Recognition (CVPR)}, 2013.

\bibitem[Yang et~al.(2019)Yang, Mei, Xu, Wei, Yin, and Lau]{yang2019where}
Xin Yang, Haiyang Mei, Ke Xu, Xiaopeng Wei, Baocai Yin, and Rynson W.~H. Lau.
\newblock Where is my mirror?
\newblock In \emph{Proceedings of the IEEE/CVF International Conference on Computer Vision (ICCV)}, 2019.

\bibitem[Yang et~al.(2024)Yang, Jiang, Hou, Zhang, Chen, and Li]{yang2024multi}
Yuqi Yang, Peng-Tao Jiang, Qibin Hou, Hao Zhang, Jinwei Chen, and Bo Li.
\newblock Multi-task dense prediction via mixture of low-rank experts.
\newblock In \emph{Proceedings of the IEEE/CVF Conference on Computer Vision and Pattern Recognition (CVPR)}, pages 27927--27937, 2024.

\bibitem[Yang et~al.(2026)Yang, Tian, Zhang, and Cao]{yang2026discover}
Yilong Yang, Jianxin Tian, Shengchuan Zhang, and Liujuan Cao.
\newblock Discover, segment, and select: A progressive mechanism for zero-shot camouflaged object segmentation.
\newblock In \emph{Proceedings of the IEEE/CVF Conference on Computer Vision and Pattern Recognition (CVPR)}, pages 34745--34754, 2026.

\bibitem[Ye et~al.(2025)Ye, Chen, Zhang, Lin, and Cao]{ye2025escnet}
Sheng Ye, Xin Chen, Yan Zhang, Xianming Lin, and Liujuan Cao.
\newblock {ESCNet}: Edge-semantic collaborative network for camouflaged object detection.
\newblock In \emph{Proceedings of the IEEE/CVF International Conference on Computer Vision (ICCV)}, pages 20053--20063, 2025.

\bibitem[You and Wu(2025)]{you2025segr1}
Zuyao You and Zuxuan Wu.
\newblock {Seg-R1}: Segmentation can be surprisingly simple with reinforcement learning.
\newblock \emph{arXiv preprint arXiv:2506.22624}, 2025.

\bibitem[You et~al.(2025)You, Kong, Meng, and Wu]{you2025focus}
Zuyao You, Lingyu Kong, Lingchen Meng, and Zuxuan Wu.
\newblock {FOCUS}: Towards universal foreground segmentation.
\newblock In \emph{Proceedings of the AAAI Conference on Artificial Intelligence (AAAI)}, pages 9580--9588, 2025.

\bibitem[You et~al.(2026)You, Wu, and Jiang]{you2026learning}
Zuyao You, Zuxuan Wu, and Yu-Gang Jiang.
\newblock Learning accurate segmentation purely from self-supervision.
\newblock In \emph{Proceedings of the European Conference on Computer Vision (ECCV)}, pages 477--494, 2026.

\bibitem[Yuan et~al.(2026)Yuan, Li, Zhang, Sun, Huang, Xu, Ji, Tong, Qi, Feng, and Yang]{yuan2026sa2va}
Haobo Yuan, Xiangtai Li, Tao Zhang, Yueyi Sun, Zilong Huang, Shilin Xu, Shunping Ji, Yunhai Tong, Lu Qi, Jiashi Feng, and Ming-Hsuan Yang.
\newblock {Sa2VA}: Marrying {SAM2} with {MLLM} for dense grounded understanding of images and videos.
\newblock \emph{IEEE Transactions on Pattern Analysis and Machine Intelligence}, 2026.

\bibitem[Yuan et~al.(2023)Yuan, Gao, and Tan]{yuan2023m3net}
Yao Yuan, Pan Gao, and XiaoYang Tan.
\newblock {M$^3$Net}: Multilevel, mixed and multistage attention network for salient object detection.
\newblock \emph{arXiv preprint arXiv:2309.08365}, 2023.

\bibitem[Yuan et~al.(2025)Yuan, Gao, Dai, Qin, and Xiang]{yuan2025uncertainty}
Yao Yuan, Pan Gao, Qun Dai, Jie Qin, and Wei Xiang.
\newblock Uncertainty-guided refinement for fine-grained salient object detection.
\newblock \emph{IEEE Transactions on Image Processing}, 34:\penalty0 2301--2314, 2025.

\bibitem[Yun and Lin(2024)]{yun2024towards}
Yi~Ke Yun and Weisi Lin.
\newblock Towards a complete and detail-preserved salient object detection.
\newblock \emph{IEEE Transactions on Multimedia}, 26:\penalty0 4667--4680, 2024.

\bibitem[Zha et~al.(2026)Zha, Wang, Li, Dong, Wang, and Yang]{zha2026seeing}
Mingfeng Zha, Guoqing Wang, Tianyu Li, Wei Dong, Peng Wang, and Yang Yang.
\newblock Seeing beyond illusion: Generalized and efficient mirror detection.
\newblock In \emph{Proceedings of the AAAI Conference on Artificial Intelligence (AAAI)}, pages 12331--12339, 2026.

\bibitem[Zhang et~al.(2023)Zhang, Rao, and Agrawala]{zhang2023adding}
Lvmin Zhang, Anyi Rao, and Maneesh Agrawala.
\newblock Adding conditional control to text-to-image diffusion models.
\newblock In \emph{Proceedings of the IEEE/CVF International Conference on Computer Vision (ICCV)}, pages 3836--3847, 2023.

\bibitem[Zhang et~al.(2025)Zhang, Ping, Li, Bing, Sun, and Knoll]{zhang2025dualgazenet}
Yu Zhang, Haoan Ping, Yuchen Li, Zhenshan Bing, Fuchun Sun, and Alois Knoll.
\newblock {DualGazeNet}: A biologically inspired dual-gaze query network for salient object detection.
\newblock \emph{arXiv preprint arXiv:2511.18865}, 2025.

\bibitem[Zhao et~al.(2026{\natexlab{a}})Zhao, Fu, He, Liu, Zhao, and Zhai]{zhao2026samba}
Wenzhuo Zhao, Keren Fu, Jiahao He, Xiaohong Liu, Qijun Zhao, and Guangtao Zhai.
\newblock {Samba+}: General and accurate salient object detection via a more unified {Mamba}-based framework.
\newblock \emph{arXiv preprint arXiv:2602.01593}, 2026{\natexlab{a}}.

\bibitem[Zhao et~al.(2026{\natexlab{b}})Zhao, Li, Mao, Xian, Jiang, Gao, Fu, Zhao, and Cheng]{zhao2026renaissance}
Wenzhuo Zhao, Xiuzhi Li, Zhongkuan Mao, Ronghao Xian, Yao Jiang, Zhao Gao, Keren Fu, Qijun Zhao, and Jian Cheng.
\newblock Is it time for the renaissance of salient object detection in the era of {MLLMs}?
\newblock \emph{arXiv preprint arXiv:2607.29222}, 2026{\natexlab{b}}.

\bibitem[Zhao et~al.(2024{\natexlab{a}})Zhao, Pang, Ji, Sheng, Zuo, Zhang, and Lu]{zhao2024spider}
Xiaoqi Zhao, Youwei Pang, Wei Ji, Baicheng Sheng, Jiaming Zuo, Lihe Zhang, and Huchuan Lu.
\newblock Spider: A unified framework for context-dependent concept segmentation.
\newblock In \emph{Proceedings of the International Conference on Machine Learning (ICML)}, pages 60906--60926, 2024{\natexlab{a}}.

\bibitem[Zhao et~al.(2024{\natexlab{b}})Zhao, Pang, Zhang, Lu, and Zhang]{zhao2024towards}
Xiaoqi Zhao, Youwei Pang, Lihe Zhang, Huchuan Lu, and Lei Zhang.
\newblock Towards diverse binary segmentation via a simple yet general gated network.
\newblock \emph{International Journal of Computer Vision}, 132\penalty0 (10):\penalty0 4157--4234, 2024{\natexlab{b}}.

\bibitem[Zhao et~al.(2026{\natexlab{c}})Zhao, Pang, Zuo, Ji, Zhou, Fan, Cao, Zhang, Liu, Lu, Lin, Tao, and Zhao]{zhao2026conceptseg}
Yuan Zhao, Youwei Pang, Jiaming Zuo, Wei Ji, Kailai Zhou, Bin Fan, Yunkang Cao, Lihe Zhang, Xiaofeng Liu, Huchuan Lu, Weisi Lin, Dacheng Tao, and Xiaoqi Zhao.
\newblock {ConceptSeg-R1}: Segment any concept via meta-reinforcement learning.
\newblock \emph{arXiv preprint arXiv:2605.20385}, 2026{\natexlab{c}}.

\bibitem[Zheng et~al.(2024)Zheng, Gao, Fan, Liu, Laaksonen, Ouyang, and Sebe]{zheng2024bilateral}
Peng Zheng, Dehong Gao, Deng-Ping Fan, Li Liu, Jorma Laaksonen, Wanli Ouyang, and Nicu Sebe.
\newblock Bilateral reference for high-resolution dichotomous image segmentation.
\newblock \emph{CAAI Artificial Intelligence Research}, 3:\penalty0 9150038, 2024.

\bibitem[Zhou et~al.(2017)Zhou, Zhao, Puig, Fidler, Barriuso, and Torralba]{zhou2017scene}
Bolei Zhou, Hang Zhao, Xavier Puig, Sanja Fidler, Adela Barriuso, and Antonio Torralba.
\newblock Scene parsing through {ADE20K} dataset.
\newblock In \emph{Proceedings of the IEEE Conference on Computer Vision and Pattern Recognition (CVPR)}, 2017.

\bibitem[Zhou et~al.(2023)Zhou, Huang, Pu, Guan, Huang, and Ling]{zhou2023treasure}
Caixia Zhou, Yaping Huang, Mengyang Pu, Qingji Guan, Li Huang, and Haibin Ling.
\newblock The treasure beneath multiple annotations: An uncertainty-aware edge detector.
\newblock In \emph{Proceedings of the IEEE/CVF Conference on Computer Vision and Pattern Recognition (CVPR)}, pages 15507--15517, 2023.

\bibitem[Zhou et~al.(2024)Zhou, Huang, Pu, Guan, Deng, and Ling]{zhou2024muge}
Caixia Zhou, Yaping Huang, Mengyang Pu, Qingji Guan, Ruoxi Deng, and Haibin Ling.
\newblock {MuGE}: Multiple granularity edge detection.
\newblock In \emph{Proceedings of the IEEE/CVF Conference on Computer Vision and Pattern Recognition (CVPR)}, pages 25952--25962, 2024.

\bibitem[Zhou et~al.(2025)Zhou, Li, Zhong, Huang, Pei, Li, and Tang]{zhou2025rethinking}
Zhangjun Zhou, Yiping Li, Chunlin Zhong, Jianuo Huang, Jialun Pei, Hua Li, and He Tang.
\newblock Rethinking detecting salient and camouflaged objects in unconstrained scenes.
\newblock In \emph{Proceedings of the IEEE/CVF International Conference on Computer Vision (ICCV)}, pages 22372--22382, 2025.

\bibitem[Zhuge et~al.(2023)Zhuge, Fan, Liu, Zhang, Xu, and Shao]{zhuge2022salient}
Mingchen Zhuge, Deng-Ping Fan, Nian Liu, Dingwen Zhang, Dong Xu, and Ling Shao.
\newblock Salient object detection via integrity learning.
\newblock \emph{IEEE Transactions on Pattern Analysis and Machine Intelligence}, 45\penalty0 (3):\penalty0 3738--3752, 2023.

\end{thebibliography}
}

\clearpage
\appendix
\section{Training details}
\label{sec:app-train}

We supervise $\sigma(S)$, resized to the label, with one loss per task.
Region tasks use binary cross-entropy (BCE) plus an IoU loss, and shadows use class-balanced BCE plus IoU.
Edges use the class-balanced BCE of RCF~\cite{liu2017richer}, which ignores pixels that some but fewer than half of the annotators mark.
Skeletons use class-balanced BCE~\cite{xie2015holistically} on targets dilated by one pixel.
The dense loss averages the per-task means, and every task contributes the same number of images per step.

AdamW updates the fusion encoder, the semantic path and the presence decoder with a learning rate of $8\times10^{-5}$ and weight decay 0.1.
The LoRA branch uses a learning rate of $2.5\times10^{-4}$ and the convention vectors $8\times10^{-4}$.
Gradients are clipped at 0.1, and the schedule has 300 warm-up steps and a cosine decay.
A step holds 192 images across all streams.
The joint presence module has its own optimizer, clipping and data stream.
The other modules thus train as in the base model, up to GPU nondeterminism.

\paragraph{Statistics.}
Each model is trained once with one seed.
Comparisons between our runs pair the images of a test set and resample them with a bootstrap.
A count of sets on which one run scores higher or lower, as in \cref{sec:exp-path,sec:exp-ablation}, includes a set only if its paired 95\% interval excludes zero.
A set whose interval contains zero counts as level.
Comparisons with published results use the printed values.

\section{Comparison with task-specific models}
\label{sec:app-compare}

The main text compares promptable and unified models, and this section lists task-specific models next to them.
\Cref{tab:rank-sod,tab:rank-cod,tab:rank-shadow,tab:rank-trans,tab:rank-mirror,tab:rank-edge,tab:rank-skel} rank the compared methods, one table per task.
A table lists only the methods that report every test set of its task, and every ranked method has released code, weights or prediction maps.
Both SAM~3 rows of these tables use the instance map of \cref{sec:method-prelim}.
Each test set ranks them, tied values share their mean rank, and a method's rank is its mean over the sets.
For salient and camouflaged objects, the S-measure sets the rank and MAE is shown beside it.
All results are taken from the original papers or reproduced by us, on the full test sets.
Most rows are the values printed in each method's own paper.
For salient and camouflaged objects, the PCM-Bench scorer re-scores Spider, DFI and VSCode from their released prediction maps or weights, and ConceptSeg-R1 is our run of its released weights.
For the other five tasks, it re-scores Spider on every set, GateNet on SBU, ISTD, Trans10K and MSD and DFI on BSDS500 and SK-LARGE, and ConceptSeg-R1 is again our run.
The SAM~3 rows and UNVEIL are our runs.
Single-task methods train on the standard training split of their task.
On the other five tasks, they also test at a single scale without CRF.
Retrained on our training data with its released code and settings, Spider reaches a seven-task mean of 79.9 (\cref{tab:scorecard}).
It takes 64 test prompts per task from the task's training images.
For edges, these are the 28,800 HED-BSDS images, while its released-weights row uses the 300 BSDS500 training and validation images.

\begin{table}[b]
\centering
\scriptsize
\setlength{\tabcolsep}{1.8pt}
\begin{tabular}{@{}lccccccc@{}}
\toprule
 & \multicolumn{2}{c}{CAMO} & \multicolumn{2}{c}{COD10K} & \multicolumn{2}{c}{NC4K} & Rank \\
\cmidrule(lr){2-3}\cmidrule(lr){4-5}\cmidrule(lr){6-7}
Method & $S_\alpha$ & $M$\,$\downarrow$ & $S_\alpha$ & $M$\,$\downarrow$ & $S_\alpha$ & $M$\,$\downarrow$ & $\downarrow$ \\
\midrule
UNVEIL (ours) & 0.918 & 0.030 & \textbf{0.933} & \textbf{0.013} & 0.927 & 0.023 & \textbf{1.7} \\
DepthSAM~\cite{han2026beyond} & \textbf{0.919} & 0.028 & 0.920 & 0.014 & \textbf{0.929} & 0.021 & 2.0 \\
SAM2-UNeXt~\cite{xiong2025sam2unext} & 0.903 & 0.036 & 0.924 & \textbf{0.013} & 0.923 & 0.022 & 4.3 \\
FOCUS~\cite{you2025focus} & 0.912 & \textbf{0.025} & 0.910 & \textbf{0.013} & 0.915 & \textbf{0.020} & 4.7 \\
NUN~\cite{he2026nested} & 0.911 & 0.028 & 0.905 & 0.014 & 0.920 & \textbf{0.020} & 5.7 \\
BiRefNet~\cite{zheng2024bilateral} & 0.904 & 0.030 & 0.913 & 0.014 & 0.914 & 0.023 & 5.8 \\
BED-SAM2~\cite{rust2026bedsam2} & 0.898 & 0.039 & 0.922 & 0.016 & 0.912 & 0.028 & 7.0 \\
DGA-Net~\cite{li2026dganet} & 0.906 & 0.033 & 0.903 & 0.018 & 0.911 & 0.026 & 8.2 \\
C3Net~\cite{jan2025c3net} & 0.904 & 0.031 & 0.898 & 0.016 & 0.913 & 0.022 & 8.3 \\
CODiff~\cite{fu2026codiff} & 0.894 & 0.033 & 0.896 & 0.014 & 0.911 & \textbf{0.020} & 11.2 \\
ARM~\cite{chen2025enhancing} & 0.887 & 0.046 & 0.909 & 0.020 & 0.906 & 0.033 & 11.7 \\
MSRNet~\cite{alghamdi2025msrnet} & 0.888 & 0.040 & 0.907 & 0.016 & 0.905 & 0.026 & 11.8 \\
DSU-Net~\cite{xu2025dsunet} & 0.889 & 0.041 & 0.891 & 0.019 & 0.909 & 0.026 & 13.0 \\
SAM~3 full fine-tuning & 0.901 & 0.033 & 0.886 & 0.022 & 0.901 & 0.029 & 14.3 \\
ZoomNeXt~\cite{pang2024zoomnext} & 0.888 & 0.040 & 0.898 & 0.017 & 0.900 & 0.028 & 14.7 \\
ARNet~\cite{wang2025assisted} & 0.890 & 0.042 & 0.883 & 0.021 & 0.899 & 0.029 & 16.7 \\
SAM2-UNet~\cite{xiong2026sam2unet} & 0.884 & 0.042 & 0.880 & 0.021 & 0.901 & 0.029 & 18.0 \\
Samba+~\cite{zhao2026samba} & 0.882 & 0.042 & 0.886 & 0.019 & 0.898 & 0.029 & 18.5 \\
ConceptSeg-R1~\cite{zhao2026conceptseg} & 0.849 & 0.055 & 0.895 & 0.021 & 0.902 & 0.027 & 18.8 \\
Selfment~\cite{you2026learning} & 0.869 & 0.060 & 0.873 & 0.033 & 0.902 & 0.033 & 20.3 \\
CamoDiffusion~\cite{chen2024camodiffusion} & 0.879 & 0.042 & 0.880 & 0.020 & 0.892 & 0.029 & 20.7 \\
RUN~\cite{he2025run} & 0.877 & 0.045 & 0.878 & 0.021 & 0.892 & 0.030 & 21.5 \\
ESCNet~\cite{ye2025escnet} & 0.871 & 0.044 & 0.873 & 0.021 & 0.892 & 0.028 & 22.7 \\
Spider~\cite{zhao2024spider} & 0.868 & 0.051 & 0.871 & 0.025 & 0.892 & 0.032 & 23.8 \\
VSCode~\cite{luo2024vscode} & 0.873 & 0.046 & 0.869 & 0.024 & 0.891 & 0.032 & 24.2 \\
DSS~\cite{yang2026discover} & 0.808 & 0.078 & 0.887 & 0.022 & 0.891 & 0.031 & 24.2 \\
FSPNet~\cite{huang2023feature} & 0.856 & 0.050 & 0.851 & 0.026 & 0.879 & 0.035 & 27.0 \\
SENet (joint, P2)~\cite{hao2025simple} & 0.861 & 0.046 & 0.849 & 0.027 & 0.878 & 0.034 & 27.3 \\
SENet (joint, P1)~\cite{hao2025simple} & 0.853 & 0.052 & 0.841 & 0.032 & 0.874 & 0.039 & 29.2 \\
GateNet~\cite{zhao2024towards} & 0.829 & 0.069 & 0.846 & 0.028 & 0.869 & 0.040 & 30.0 \\
CDP~\cite{liu2025seamless} & 0.798 & 0.072 & 0.791 & 0.036 & 0.880 & 0.051 & 30.3 \\
HVPNet~\cite{xu2027hvpnet} & 0.834 & 0.060 & 0.841 & 0.027 & 0.865 & 0.038 & 30.5 \\
SAM~3 (frozen) & 0.519 & 0.174 & 0.503 & 0.092 & 0.517 & 0.147 & 33.0 \\
\bottomrule
\end{tabular}
\caption{\textbf{Camouflaged objects.}
Notation as in \cref{tab:rank-sod}.}
\label{tab:rank-cod}
\end{table}

\begin{table*}[t]
\begin{minipage}[t]{0.63\textwidth}
\vspace{0pt}
\centering
\scriptsize
\setlength{\tabcolsep}{2.6pt}
\begin{tabular}{@{}lccccccccccc@{}}
\toprule
 & \multicolumn{2}{c}{DUTS-TE} & \multicolumn{2}{c}{ECSSD} & \multicolumn{2}{c}{HKU-IS} & \multicolumn{2}{c}{DUT-OMRON} & \multicolumn{2}{c}{PASCAL-S} & Rank \\
\cmidrule(lr){2-3}\cmidrule(lr){4-5}\cmidrule(lr){6-7}\cmidrule(lr){8-9}\cmidrule(lr){10-11}
Method & $S_\alpha$ & $M$\,$\downarrow$ & $S_\alpha$ & $M$\,$\downarrow$ & $S_\alpha$ & $M$\,$\downarrow$ & $S_\alpha$ & $M$\,$\downarrow$ & $S_\alpha$ & $M$\,$\downarrow$ & $\downarrow$ \\
\midrule
DualGazeNet~\cite{zhang2025dualgazenet} & 0.939 & 0.019 & \textbf{0.956} & 0.018 & 0.944 & 0.019 & 0.894 & \textbf{0.034} & 0.902 & 0.041 & \textbf{4.0} \\
UNVEIL (ours) & \textbf{0.947} & \textbf{0.017} & \textbf{0.956} & 0.016 & 0.944 & 0.018 & 0.901 & \textbf{0.034} & 0.897 & 0.042 & 4.2 \\
BED-SAM2~\cite{rust2026bedsam2} & 0.943 & 0.018 & 0.952 & 0.021 & \textbf{0.947} & 0.018 & 0.891 & 0.035 & 0.898 & 0.043 & 4.9 \\
LFNet~\cite{chen2026liquid} & 0.936 & -- & 0.954 & -- & 0.946 & -- & \textbf{0.904} & -- & 0.896 & -- & 5.0 \\
Samba+~\cite{zhao2026samba} & 0.936 & -- & 0.955 & -- & \textbf{0.947} & -- & 0.893 & -- & 0.896 & -- & 5.3 \\
Samba (VMamba-B)~\cite{zhao2026samba} & 0.934 & 0.020 & \textbf{0.956} & 0.018 & \textbf{0.947} & 0.018 & 0.890 & 0.037 & 0.896 & 0.048 & 6.0 \\
DSU-Net~\cite{xu2025dsunet} & 0.934 & 0.020 & 0.953 & 0.018 & 0.945 & 0.019 & 0.893 & \textbf{0.034} & 0.898 & 0.042 & 6.4 \\
SAM3-UNet~\cite{xiong2025sam3unet} & 0.936 & 0.019 & 0.950 & 0.019 & 0.939 & 0.020 & 0.895 & \textbf{0.034} & \textbf{0.904} & 0.038 & 8.7 \\
Samba~\cite{he2025samba} & 0.932 & 0.020 & 0.953 & 0.019 & 0.945 & 0.018 & 0.889 & 0.037 & 0.892 & 0.047 & 9.4 \\
ConceptSeg-R1~\cite{zhao2026conceptseg} & 0.937 & 0.018 & 0.952 & \textbf{0.015} & 0.935 & 0.020 & 0.877 & 0.044 & 0.900 & 0.037 & 11.0 \\
SAM2-UNet~\cite{xiong2026sam2unet} & 0.934 & 0.020 & 0.950 & 0.020 & 0.941 & 0.019 & 0.884 & 0.039 & 0.894 & 0.043 & 11.4 \\
UGRAN~\cite{yuan2025uncertainty} & 0.931 & 0.022 & 0.950 & 0.020 & 0.945 & 0.019 & 0.878 & 0.045 & 0.892 & 0.046 & 11.9 \\
InSPyReNet~\cite{kim2022revisiting} & 0.931 & 0.024 & 0.949 & 0.023 & 0.944 & 0.021 & 0.875 & 0.045 & 0.893 & 0.048 & 14.3 \\
VisionLLM~v2~\cite{wu2024visionllm} & 0.921 & 0.024 & 0.950 & 0.018 & 0.941 & \textbf{0.016} & 0.882 & 0.041 & 0.892 & 0.044 & 14.8 \\
SENet (joint, P2)~\cite{hao2025simple} & 0.929 & 0.021 & 0.950 & 0.019 & 0.940 & 0.019 & 0.874 & 0.048 & 0.891 & 0.046 & 16.6 \\
SENet~\cite{hao2025simple} & 0.926 & 0.022 & 0.949 & 0.020 & 0.941 & 0.019 & 0.876 & 0.040 & 0.890 & 0.046 & 17.0 \\
UniSOD~\cite{wang2025unified} & 0.925 & 0.021 & 0.950 & 0.017 & 0.940 & 0.018 & 0.876 & 0.037 & 0.889 & 0.044 & 17.5 \\
SENet (joint, P1)~\cite{hao2025simple} & 0.932 & 0.020 & 0.946 & 0.022 & 0.939 & 0.020 & 0.878 & 0.039 & 0.887 & 0.047 & 17.6 \\
M3Net~\cite{yuan2023m3net} & 0.927 & 0.024 & 0.948 & 0.021 & 0.943 & 0.019 & 0.872 & 0.045 & 0.889 & 0.047 & 17.9 \\
VST++~\cite{liu2024vst} & 0.923 & 0.024 & 0.950 & 0.021 & 0.941 & 0.021 & 0.871 & 0.044 & 0.889 & 0.056 & 18.0 \\
VSCode~\cite{luo2024vscode} & 0.926 & 0.024 & 0.949 & 0.021 & 0.940 & 0.021 & 0.877 & 0.043 & 0.886 & 0.050 & 18.4 \\
FOCUS~\cite{you2025focus} & 0.929 & 0.019 & 0.943 & 0.018 & 0.935 & 0.018 & 0.868 & 0.045 & 0.898 & \textbf{0.036} & 18.7 \\
MDSAM~\cite{gao2024multi} & 0.920 & 0.024 & 0.948 & 0.021 & 0.941 & 0.019 & 0.878 & 0.039 & 0.882 & 0.052 & 18.9 \\
EVPv2 (ViT-SAM)~\cite{liu2026explicit} & 0.915 & 0.026 & 0.944 & 0.022 & 0.937 & 0.020 & 0.874 & 0.040 & 0.877 & 0.054 & 24.1 \\
ICON-S~\cite{zhuge2022salient} & 0.917 & 0.025 & 0.941 & 0.023 & 0.935 & 0.022 & 0.869 & 0.043 & 0.885 & 0.048 & 24.2 \\
EVPv2 (MiT-B4)~\cite{liu2026explicit} & 0.915 & 0.027 & 0.935 & 0.028 & 0.932 & 0.023 & 0.862 & 0.047 & 0.879 & 0.053 & 26.8 \\
SAM~3 full fine-tuning & 0.915 & 0.031 & 0.939 & 0.024 & 0.932 & 0.023 & 0.843 & 0.064 & 0.880 & 0.050 & 27.5 \\
GateNet~\cite{zhao2024towards} & 0.906 & 0.030 & 0.941 & 0.026 & 0.931 & 0.025 & 0.847 & 0.050 & 0.875 & 0.055 & 28.7 \\
SelfReformer~\cite{yun2024towards} & 0.911 & 0.026 & 0.935 & 0.027 & 0.930 & 0.024 & 0.856 & 0.041 & 0.874 & 0.050 & 28.9 \\
MENet~\cite{wang2023pixels} & 0.905 & 0.028 & 0.928 & 0.031 & 0.927 & 0.023 & 0.850 & 0.045 & 0.872 & 0.054 & 30.2 \\
Spider~\cite{zhao2024spider} & 0.917 & 0.027 & 0.920 & 0.039 & 0.899 & 0.043 & 0.857 & 0.052 & 0.815 & 0.095 & 30.7 \\
RAG-SEG~\cite{liu2025first} & 0.897 & 0.032 & 0.927 & 0.028 & 0.917 & 0.026 & 0.804 & 0.065 & 0.878 & 0.045 & 31.6 \\
EDN~\cite{wu2022edn} & 0.892 & 0.035 & 0.927 & 0.033 & 0.924 & 0.027 & 0.849 & 0.050 & 0.865 & 0.062 & 31.7 \\
DFI~\cite{liu2020dynamic} & 0.887 & 0.039 & 0.927 & 0.035 & 0.920 & 0.031 & 0.840 & 0.055 & 0.865 & 0.065 & 32.7 \\
SAM~3 (frozen) & 0.437 & 0.151 & 0.405 & 0.228 & 0.441 & 0.180 & 0.440 & 0.151 & 0.402 & 0.239 & 35.0 \\
\bottomrule
\end{tabular}
\caption{\textbf{Salient objects.}
Methods that report every test set, ordered by the mean rank of $S_\alpha$. $M$: MAE. --: not reported.
Bold: best.}
\label{tab:rank-sod}
\end{minipage}\hfill
\begin{minipage}[t]{0.35\textwidth}
\vspace{0pt}
\centering
\footnotesize
\setlength{\tabcolsep}{4pt}
\begin{tabular}{@{}lccc@{}}
\toprule
 & SBU & ISTD & Rank \\
\cmidrule(lr){2-2}\cmidrule(lr){3-3}
Method & BER\,$\downarrow$ & BER\,$\downarrow$ & $\downarrow$ \\
\midrule
ShadowAdapter~\cite{jie2025shadowadapter} & 2.75 & 0.86 & \textbf{2.0} \\
UNVEIL (ours) & 2.83 & \textbf{0.84} & \textbf{2.0} \\
SVL~\cite{wang2026revisiting} & \textbf{2.46} & 1.04 & 2.3 \\
GateNet~\cite{zhao2024towards} & 4.06 & 1.04 & 4.8 \\
Spider~\cite{zhao2024spider} & 3.11 & 3.94 & 5.0 \\
SAM~3 full fine-tuning & 4.02 & 1.51 & 5.0 \\
ConceptSeg-R1~\cite{zhao2026conceptseg} & 11.92 & 7.77 & 7.0 \\
SAM~3 (frozen) & 12.62 & 10.25 & 8.0 \\
\bottomrule
\end{tabular}
\caption{\textbf{Shadows.}
Notation as in \cref{tab:rank-sod}.}
\label{tab:rank-shadow}

\bigskip
\begin{tabular}{@{}lcc@{}}
\toprule
 & MSD & Rank \\
\cmidrule(lr){2-2}
Method & IoU & $\downarrow$ \\
\midrule
UNVEIL (ours) & \textbf{0.964} & \textbf{1.0} \\
SAM~3 full fine-tuning & 0.960 & 2.0 \\
SAM3-UNet~\cite{xiong2025sam3unet} & 0.943 & 3.0 \\
SAM2-UNet~\cite{xiong2026sam2unet} & 0.918 & 4.0 \\
MirrorSAM~\cite{zha2026seeing} & 0.913 & 5.0 \\
GlassWizard~\cite{li2025glasswizard} & 0.911 & 6.0 \\
M2SD~\cite{shao2025mirror} & 0.871 & 7.0 \\
SATNet~\cite{huang2023symmetry} & 0.854 & 8.0 \\
ConceptSeg-R1~\cite{zhao2026conceptseg} & 0.831 & 9.0 \\
HetNet~\cite{he2023efficient} & 0.828 & 10.0 \\
CSFwinformer~\cite{xie2024csfwinformer} & 0.821 & 11.0 \\
GateNet~\cite{zhao2024towards} & 0.810 & 12.0 \\
SAM~3 (frozen) & 0.692 & 13.0 \\
Spider~\cite{zhao2024spider} & 0.578 & 14.0 \\
\bottomrule
\end{tabular}
\caption{\textbf{Mirrors.}
Notation as in \cref{tab:rank-sod}.}
\label{tab:rank-mirror}
\end{minipage}
\end{table*}

By mean rank, UNVEIL is at position 2 of 35 on salient objects and 1 of 33 on camouflaged objects.
It shares position 1 of 8 on shadows with ShadowAdapter~\cite{jie2025shadowadapter}.
It is at position 1 of 6 on transparent objects, 1 of 14 on mirrors, 1 of 18 on edges and 1 of 8 on skeletons.
ConceptSeg-R1 hands almost every shadow and transparent image to SAM~3, and its BER on SBU is 11.92.

For salient and camouflaged objects, we also re-score every released prediction map with the scorer of our main tables.
Two methods have no released maps, and we ran their official code and weights.
The paired comparison takes the images that both models cover and resamples them 1,000 times.
Over 100 pairs of method and test set, the S-measure favors UNVEIL on 85 and the other method on 2.
The counts are 81 and 4 for the weighted F-measure, 74 and 7 for the mean E-measure, and 70 and 5 for MAE, with the remaining pairs level.

\Cref{tab:extrasets} scores the rows of \cref{tab:scorecard} on every test set of six tasks.
After the sets of PCM-Bench come UCF for shadows, GSD for transparent objects, PMD for mirrors, CHAMELEON for camouflaged objects, BIPEDv2 for edges and WH-SYMMAX for skeletons.
The metrics and thresholds are those of \cref{sec:exp-setup}.
UNVEIL has the best value on every set.

\section{Qualitative results}
\label{sec:app-qual}

\begin{figure*}[t]
\centering
\includegraphics[width=\textwidth]{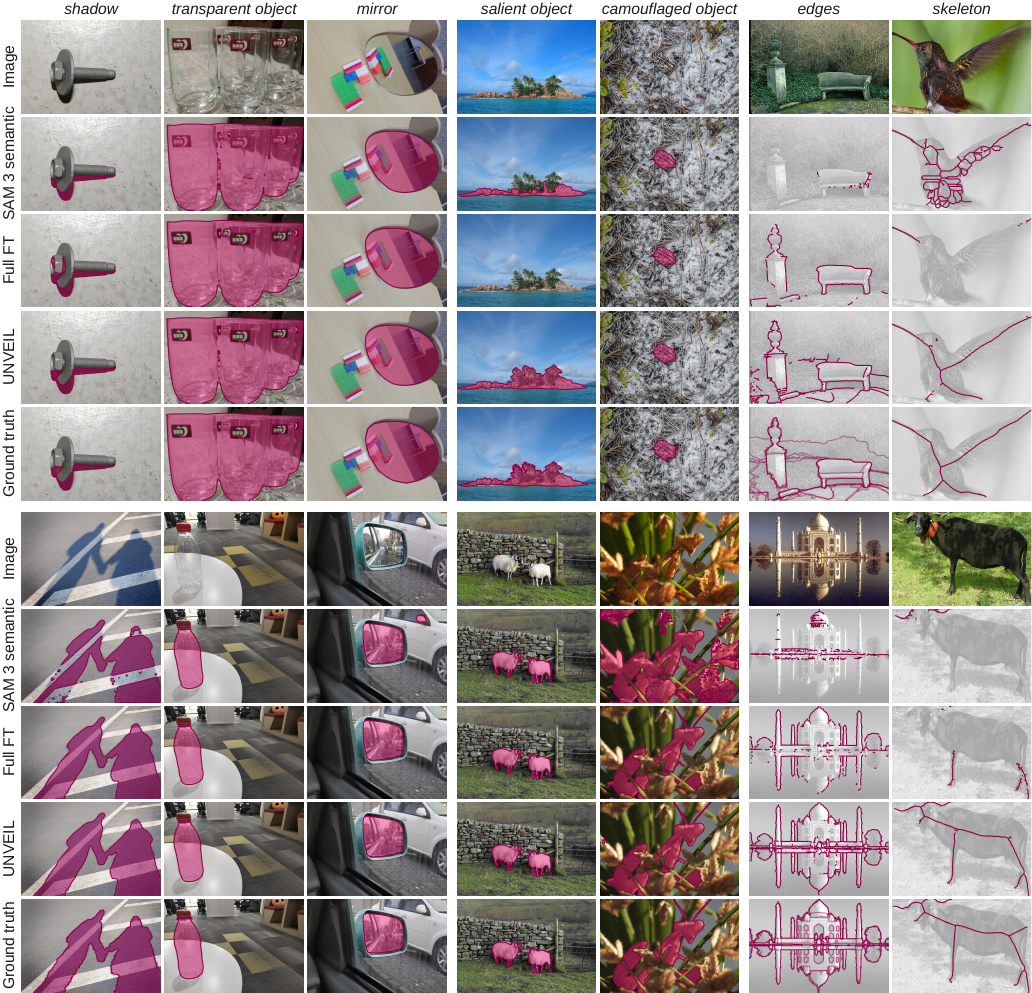}
\caption{\textbf{Random test images.}
Two images per prompt, drawn at random from the full test sets (\cref{sec:app-qual}).
Rows show the image, frozen SAM~3's semantic path (SAM~3 semantic), SAM~3's official full fine-tuning on our data (Full FT), UNVEIL and the ground truth.}
\label{fig:qual}
\end{figure*}

\Cref{fig:qual} shows two test images per prompt, drawn at random with a fixed seed from the full test sets of PCM-Bench.
The draw leaves out images without a target or with a tiny one, and images that do not fit the 3:2 cells of \cref{fig:qual-main}.
UNVEIL queries SYM-PASCAL with the prompt ``symmetry axis'', so skeletons come from SK-LARGE.
The SAM~3 row shows the unmodified semantic path with the prompts of UNVEIL, as in \cref{sec:veto}.
Full fine-tuning and UNVEIL binarize shadows, transparent objects and mirrors at the thresholds from the training splits, and all other region maps at 0.5.
Edges and skeletons appear as thinned lines at each model's ODS threshold.

\begin{table}[t]
\centering
\footnotesize
\setlength{\tabcolsep}{4pt}
\begin{tabular}{@{}lcc@{}}
\toprule
 & Trans10K & Rank \\
\cmidrule(lr){2-2}
Method & IoU & $\downarrow$ \\
\midrule
UNVEIL (ours) & \textbf{0.956} & \textbf{1.0} \\
SAM~3 full fine-tuning & 0.951 & 2.0 \\
GateNet~\cite{zhao2024towards} & 0.936 & 3.0 \\
Spider~\cite{zhao2024spider} & 0.933 & 4.0 \\
ConceptSeg-R1~\cite{zhao2026conceptseg} & 0.886 & 5.0 \\
SAM~3 (frozen) & 0.585 & 6.0 \\
\bottomrule
\end{tabular}
\caption{\textbf{Transparent objects.}
Notation as in \cref{tab:rank-sod}.}
\label{tab:rank-trans}
\end{table}

\Cref{fig:qual-main} draws from SBU, Trans10K, MSD, DUTS-TE, COD10K, BSDS500 and SK-LARGE.
Difficulty is the mean per-image score of SAM~3's full and LoRA fine-tuning.
The figure keeps the hardest tenth of each set and the hardest fifth of SK-LARGE, whose images rarely fit its 3:2 cells.
Images without a target or with a tiny one, and images that do not fit these cells, are left out.
Each example is the image closest to the median of UNVEIL's scores in this pool.
Full FT and UNVEIL are drawn as in \cref{fig:qual}.

\begin{table*}[t]
\centering
\begin{minipage}[t]{0.40\textwidth}
\vspace{0pt}
\centering
\footnotesize
\setlength{\tabcolsep}{4pt}
\begin{tabular}{@{}lcc@{}}
\toprule
 & BSDS500 & Rank \\
\cmidrule(lr){2-2}
Method & ODS & $\downarrow$ \\
\midrule
UNVEIL (ours) & \textbf{0.856} & \textbf{1.0} \\
SAUGE~\cite{liufu2025sauge} & 0.847 & 2.0 \\
EdgeNAT~\cite{jie2024edgenat} & 0.843 & 3.0 \\
NBED~\cite{li2026new} & 0.838 & 4.0 \\
EDMB~\cite{li2025edmb} & 0.837 & 5.0 \\
MuGE~\cite{zhou2024muge} & 0.831 & 6.0 \\
UAED~\cite{zhou2023treasure} & 0.829 & 7.0 \\
EDTER~\cite{pu2022edter} & 0.824 & 8.5 \\
RankED~\cite{cetinkaya2024ranked} & 0.824 & 8.5 \\
BDCN~\cite{he2019bidirectional} & 0.806 & 10.0 \\
DFI~\cite{liu2020dynamic} & 0.799 & 11.0 \\
RCF~\cite{liu2017richer} & 0.798 & 12.0 \\
PiDiNet~\cite{su2021pixel} & 0.789 & 13.0 \\
HED~\cite{xie2015holistically} & 0.788 & 14.0 \\
SAM~3 full fine-tuning & 0.706 & 15.0 \\
ConceptSeg-R1~\cite{zhao2026conceptseg} & 0.323 & 16.0 \\
Spider~\cite{zhao2024spider} & 0.216 & 17.0 \\
SAM~3 (frozen) & 0.035 & 18.0 \\
\bottomrule
\end{tabular}
\caption{\textbf{Edges.}
Notation as in \cref{tab:rank-sod}.}
\label{tab:rank-edge}
\end{minipage}\hspace{0.06\textwidth}
\begin{minipage}[t]{0.44\textwidth}
\vspace{0pt}
\centering
\footnotesize
\setlength{\tabcolsep}{4pt}
\begin{tabular}{@{}lccc@{}}
\toprule
 & SK-LARGE & SYM-PASCAL & Rank \\
\cmidrule(lr){2-2}\cmidrule(lr){3-3}
Method & ODS & ODS & $\downarrow$ \\
\midrule
UNVEIL (ours) & \textbf{0.793} & \textbf{0.647} & \textbf{1.0} \\
DeepFlux~\cite{wang2019deepflux} & 0.732 & 0.502 & 3.0 \\
AdaLSN~\cite{liu2021adaptive} & 0.786 & 0.497 & 3.0 \\
DFI~\cite{liu2020dynamic} & 0.723 & 0.511 & 3.0 \\
Spider~\cite{zhao2024spider} & 0.642 & 0.407 & 5.0 \\
ConceptSeg-R1~\cite{zhao2026conceptseg} & 0.554 & 0.348 & 6.0 \\
SAM~3 full fine-tuning & 0.317 & 0.327 & 7.0 \\
SAM~3 (frozen) & 0.233 & 0.000 & 8.0 \\
\bottomrule
\end{tabular}
\caption{\textbf{Skeletons.}
Notation as in \cref{tab:rank-sod}.}
\label{tab:rank-skel}
\end{minipage}
\end{table*}

\begin{table*}[t]
\centering
\scriptsize
\setlength{\tabcolsep}{2.2pt}
\begin{tabular}{lccccccc}
\toprule
 & \multicolumn{3}{c}{Shadow} & \multicolumn{2}{c}{Transparent} & \multicolumn{2}{c}{Mirror} \\
\cmidrule(lr){2-4}\cmidrule(lr){5-6}\cmidrule(lr){7-8}
 & SBU & ISTD & UCF & Trans10K & GSD & MSD & PMD \\
 & BER\,$\downarrow$ & BER\,$\downarrow$ & BER\,$\downarrow$ & IoU & IoU & IoU & IoU \\
\midrule
SAM~3 (frozen) & 12.62 & 10.25 & 16.55 & 0.585 & 0.234 & 0.692 & 0.644 \\
SAM~3, presence set to one & 12.39 & 10.57 & 15.61 & 0.806 & 0.703 & 0.769 & 0.697 \\
SAM~3 Agent (7B)~\cite{carion2025sam3} & 22.11 & 17.51 & 25.38 & 0.697 & 0.410 & 0.727 & 0.620 \\
\midrule
LISA-7B~\cite{lai2024lisa} & 34.99 & 31.12 & 43.24 & 0.533 & 0.171 & 0.529 & 0.462 \\
GLaMM~\cite{rasheed2024glamm} & 32.20 & 24.89 & 37.23 & 0.588 & 0.268 & 0.619 & 0.455 \\
PixelLM-7B~\cite{ren2024pixellm} & 37.38 & 33.92 & 47.69 & 0.589 & 0.302 & 0.617 & 0.558 \\
Grounded SAM~2~\cite{ren2024grounded} & 34.95 & 28.64 & 36.55 & 0.456 & 0.105 & 0.240 & 0.372 \\
Sa2VA-8B~\cite{yuan2026sa2va} & 27.81 & 20.13 & 30.68 & 0.721 & 0.428 & 0.563 & 0.389 \\
ConceptSeg-R1~\cite{zhao2026conceptseg} & 11.92 & 7.77 & 16.61 & 0.886 & 0.824 & 0.831 & 0.734 \\
\midrule
Sa2VA, fine-tuned on our data & 9.14 & 9.31 & 13.87 & 0.864 & 0.784 & 0.878 & 0.656 \\
Spider~\cite{zhao2024spider}, retrained on our data & 4.00 & 1.35 & 7.39 & 0.935 & 0.665 & 0.877 & 0.218 \\
SAM~3 full fine-tuning & 4.02 & 1.51 & 6.56 & 0.951 & 0.880 & 0.960 & 0.721 \\
SAM~3 LoRA fine-tuning & 4.40 & 1.74 & 6.86 & 0.950 & 0.867 & 0.955 & 0.702 \\
\midrule
UNVEIL (ours) & \textbf{2.83} & \textbf{0.84} & \textbf{5.98} & \textbf{0.956} & \textbf{0.887} & \textbf{0.964} & \textbf{0.752} \\
\bottomrule
\end{tabular}
\\[2pt]
\begin{tabular}{lccccccccc}
\toprule
 & \multicolumn{4}{c}{Camouflaged} & \multicolumn{2}{c}{Edges} & \multicolumn{3}{c}{Skeleton} \\
\cmidrule(lr){2-5}\cmidrule(lr){6-7}\cmidrule(lr){8-10}
 & CAMO & COD10K & NC4K & CHAMELEON & BSDS500 & BIPEDv2 & SK-LARGE & SYM-PASCAL & WH-SYMMAX \\
 & $S_\alpha$ & $S_\alpha$ & $S_\alpha$ & $S_\alpha$ & ODS & ODS & ODS & ODS & ODS \\
\midrule
SAM~3 (frozen) & 0.519 & 0.503 & 0.517 & 0.515 & 0.035 & 0.013 & 0.233 & 0.000 & 0.292 \\
SAM~3, presence set to one & 0.504 & 0.483 & 0.535 & 0.470 & 0.340 & 0.532 & 0.532 & 0.190 & 0.776 \\
SAM~3 Agent (7B)~\cite{carion2025sam3} & 0.742 & 0.806 & 0.834 & 0.811 & 0.086 & 0.087 & 0.041 & 0.109 & 0.000 \\
\midrule
LISA-7B~\cite{lai2024lisa} & 0.678 & 0.716 & 0.694 & 0.774 & 0.126 & 0.024 & 0.229 & 0.013 & 0.179 \\
GLaMM~\cite{rasheed2024glamm} & 0.746 & 0.704 & 0.698 & 0.764 & 0.214 & 0.196 & 0.389 & 0.255 & 0.450 \\
PixelLM-7B~\cite{ren2024pixellm} & 0.712 & 0.701 & 0.746 & 0.720 & 0.180 & 0.119 & 0.347 & 0.107 & 0.084 \\
Grounded SAM~2~\cite{ren2024grounded} & 0.481 & 0.570 & 0.636 & 0.453 & 0.125 & 0.134 & 0.571 & 0.439 & 0.864 \\
Sa2VA-8B~\cite{yuan2026sa2va} & 0.765 & 0.759 & 0.743 & 0.798 & 0.124 & 0.096 & 0.581 & 0.238 & 0.736 \\
ConceptSeg-R1~\cite{zhao2026conceptseg} & 0.849 & 0.895 & 0.902 & 0.906 & 0.323 & 0.376 & 0.554 & 0.348 & 0.797 \\
\midrule
Sa2VA, fine-tuned on our data & 0.869 & 0.886 & 0.893 & 0.883 & 0.086 & 0.011 & 0.118 & 0.037 & 0.317 \\
Spider~\cite{zhao2024spider}, retrained on our data & 0.878 & 0.877 & 0.895 & 0.918 & 0.483 & 0.418 & 0.759 & 0.460 & 0.834 \\
SAM~3 full fine-tuning & 0.901 & 0.886 & 0.901 & 0.931 & 0.706 & 0.326 & 0.317 & 0.327 & 0.527 \\
SAM~3 LoRA fine-tuning & 0.899 & 0.907 & 0.893 & 0.927 & 0.634 & 0.323 & 0.550 & 0.420 & 0.729 \\
\midrule
UNVEIL (ours) & \textbf{0.918} & \textbf{0.933} & \textbf{0.927} & \textbf{0.945} & \textbf{0.856} & \textbf{0.703} & \textbf{0.793} & \textbf{0.647} & \textbf{0.879} \\
\bottomrule
\end{tabular}
\caption{\textbf{Every test set of six tasks}, rows as in \cref{tab:scorecard}.
The PCM-Bench sets of each task come first.
Bold: best.}
\label{tab:extrasets}
\end{table*}

\end{document}